\documentclass[10pt,twocolumn,letterpaper]{article}

\usepackage[pagenumbers]{cvpr}            
\newcommand{\topic}[1]{%
    \par\noindent\textbf{#1} %
}

\definecolor{cvprblue}{rgb}{0.21,0.49,0.74}
\usepackage[pagebackref,breaklinks,colorlinks,allcolors=cvprblue]{hyperref}
\usepackage{tabularray,xcolor, multirow}
\usepackage{algorithm}
\usepackage{listings}
\usepackage{amsmath}
\usepackage{color, colortbl} 
\UseTblrLibrary{siunitx}

\definecolor{darkF7E0D5}{RGB}{209,154,128}
\definecolor{Gray}{gray}{0.90}
\definecolor{light}{RGB}{179, 224, 255}
\definecolor{darkblue}{RGB}{0, 112, 188}
\definecolor{newyellow}{RGB}{197, 197, 0}

\definecolor{cvprblue}{rgb}{0.21,0.49,0.74}
\usepackage[pagebackref,breaklinks,colorlinks,allcolors=cvprblue]{hyperref}

\def\paperID{24} 
\def\confName{CVPRW}
\def\confYear{2025}

\newcommand{\posval}[1]{\textcolor{teal}{#1}}
\newcommand{\negval}[1]{\textcolor{red}{#1}}

\title{Does Feasibility Matter? \\Understanding the Impact of Feasibility on Synthetic Training Data}

\author{
Yiwen Liu\textsuperscript{1} \quad\quad
Jessica Bader\textsuperscript{1,3}  \quad\quad
Jae Myung Kim\textsuperscript{2,3}\\[5pt]
\textsuperscript{1}Technical University of Munich \quad \textsuperscript{2}University of Tübingen  \quad \textsuperscript{3}Helmholtz Munich 
}

\begin{document}

\twocolumn[{%
\renewcommand\twocolumn[1][]{#1}%
\maketitle

\begin{center}
    \vspace{-1.em}
    \includegraphics[width=1.0\linewidth]{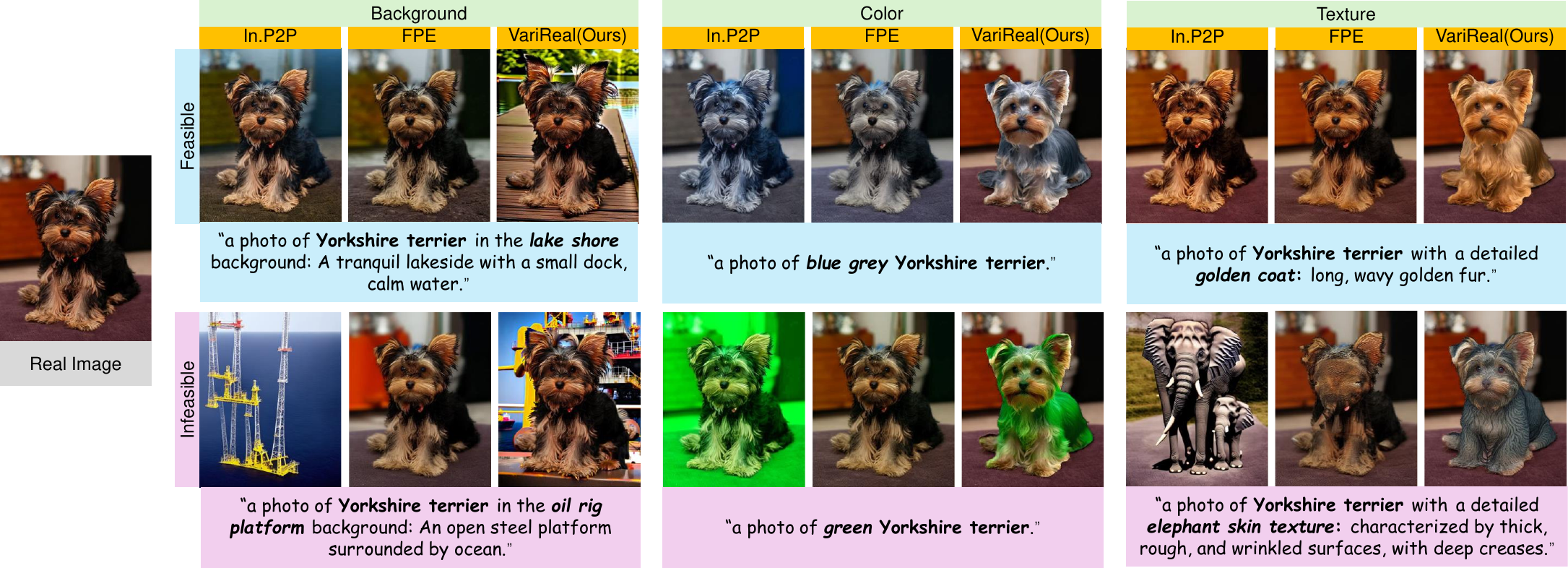}
    \vspace{-1.5em}
    \captionof{figure}{
    We propose VariReal, a pipeline for minimal-change editing of real images, enabling isolation of target attributes in three categories: background, color, and texture. We compare images generated by VariReal to those produced by prior text-guided editing methods~\cite{instructpix2pix,fpe}, examining both feasible and infeasible attributes. The editing prompts are provided below each generated image.
    }
    
    \label{fig: teaser}
   
\end{center}%
}]

\maketitle

\begin{abstract}
With the development of photorealistic diffusion models, models trained in part or fully on synthetic data achieve progressively better results. However, diffusion models still routinely generate images that would not exist in reality, such as a dog floating above the ground or with unrealistic texture artifacts. We define the concept of \textbf{feasibility} as whether attributes in a synthetic image could realistically exist in the real-world domain; synthetic images containing attributes that violate this criterion are considered \textbf{infeasible}. Intuitively, infeasible images are typically considered out-of-distribution; thus, training on such images is expected to hinder a model's ability to generalize to real-world data, and they should therefore be excluded from the training set whenever possible. However, does feasibility really matter? In this paper, we investigate whether enforcing feasibility is necessary when generating synthetic training data for CLIP-based classifiers, focusing on three target attributes: background, color, and texture. We introduce VariReal, a pipeline that minimally edits a given source image to include feasible or infeasible attributes given by the textual prompt generated by a large language model. Our experiments show that feasibility minimally affects LoRA-fine-tuned CLIP performance, with mostly less than 0.3\% difference in top-1 accuracy across three fine-grained datasets. Also, the attribute matters on whether the feasible/infeasible images adversarially influence the classification performance. Finally, mixing feasible and infeasible images in training datasets does not significantly impact performance compared to using purely feasible or infeasible datasets. Code is available at \href{https://github.com/Yiveen/SyntheticDataFeasibility}{https://github.com/Yiveen/SyntheticDataFeasibility}.

\end{abstract}

\vspace{-5mm}
\section{Introduction}
\label{sec:intro}

In recent years, large-scale pre-trained models~\cite{diffusion, sd, chatgpt, internvl,dino, sam} have significantly surpassed traditional learning approaches in various tasks. However, as the scale of training data grows, access to high-quality data has become increasingly limited~\cite{zhou2024programming}, posing challenges to further improving these large models' capabilities. With the popularity of generative models~\cite{vae,gan} like Stable Diffusion~\cite{sd}, researchers are increasingly leveraging these models to generate high-fidelity synthetic data that closely resembles real-world data, offering a solution to data scarcity~\cite{surveydpm, surveyhealth}.

Prior studies have explored synthetic data generation under a limited few-shot real image setting~\cite{yu2023diversify,kim2024datadream,sariyildiz2023fake,he2022synthetic,dunlap2023diversify,da2023diversified,hammoud2024synthclip,wu2023diffumask}. These works aim to create synthetic data that approximates the real-world data  distribution while avoiding overfitting to the limited available examples. Some studies~\cite{kim2024datadream, he2022synthetic} suggest that synthetic data can offer benefits beyond those of real data. However, the inherent randomness in the diffusion-based image generation process~\cite{diffusion, sd} can introduce domain shifts~\cite{he2022synthetic} or implausible scenarios, such as “a dog floating in the sky”~\cite{sariyildiz2023fake}, which fail to reflect realistic patterns. Such data could intuitively be perceived as out-of-distribution (OOD), potentially becoming counterproductive for downstream tasks.

Interestingly, previous studies~\cite{de2023value,geiping2022much,bengio2011deep} suggest that OOD data can positively impact downstream tasks when mixed with real data in certain proportions. A typical example is data augmentation~\cite{geiping2022much}, where some data augment methods introduce OOD data relative to the original distribution yet still provide benefits. While these advantages generally diminish as divergence from the original distribution increases~\cite{de2023value}, these findings demonstrate OOD data is not always harmful. Conversely, incorporating feasible content, which is considered in-distribution, is naturally beneficial. For instance, \citet{dunlap2023diversify} propose augmenting training data by synthesizing data with diverse feasible backgrounds and show performance gain. This raises a question: \textit{does the feasibility matter for synthetic training data?}

In this paper, we study the impact of the feasibility on synthesized data when using them as training data for the classification task. We define feasibility as whether class-specific attributes could realistically occur in the real world.
Attributes that meet this criterion are considered feasible while others are infeasible. For instance, given a Yorkshire terrier in Figure~\ref{fig: teaser}, it is likely to find it at the lake shore, while not at the oil rig platform. Therefore, we assume an image of Yorkshire terrier at the lake shore background as a feasible image, while an image of it at the oil rig platform as an infeasible image.

To generate feasible and infeasible images and study their impact of a downstream classification task, we propose VariReal, an editing pipeline with minimal change of attributes given a real image. We first generate a list of feasible and infeasible attribute names for each class by using GPT-4~\cite{gpt4}, with generated attributes further being validated through a user study. We then use a proposed image-editing pipeline based on Stable Diffusion~\cite{sd} that generates feasible (or infeasible) images given a source real image and a prompt with a feasible (or infeasible) attribute name. We then assess the impact of the feasibility of images to downstream tasks by fine-tuning CLIP-based classifiers under two conditions: synthetic-only training and mixed real-synthetic training. 

Our study of feasibility for a downstream task in three different attributes (background, color, texture) on three fine-grained datasets reveals the following insights. First, we show that changing the background regardless of feasibility brings performance gain, which loosens a restriction considered in ALIA~\cite{dunlap2023diversify} where it only uses a feasible background scenario. Second, foreground modifications, like color or texture attributes, often challenge the classifier's learning process especially when the training datasets are infeasible inputs. 

In summary, our contributions are as follows:
\begin{itemize} 
    \item We propose VariReal, an automated generation pipeline for producing minimal-change synthetic data by altering only one attribute from real images at a time. This approach can be applied out-of-the-box to any object-centric classification dataset without additional fine-tuning. 
    \item We define and generate feasible and infeasible dataset comparison pairs based on real images, covering three controlled attributes. 
    \item To explore feasible and infeasible data roles, we fine-tune CLIP with LoRA~\cite{lora}. Analyzing classification scores, we offer new insights into the impact of feasibility and the strategic use of synthetic data for enhancing downstream classification performance. 
\end{itemize}

\section{Related Work}
\label{sec:relatedwork}
\topic{Effect of out-of-distribution data.}OOD data, defined relative to in-distribution data, introduces a distribution shift between train and test data. OOD data is generally categorized into semantic and covariance shifts~\cite{yang2024generalized}; here, we focus on covariance shifts. The impact of OOD data is commonly evaluated using classification tasks~\cite{bengio2011deep, geiping2022much,de2023value,geiping2022much}. Early works \cite{bengio2011deep, geiping2022much} attributed OOD data's benefits to feature invariance and the stochasticity it adds in gradient descent, helping avoid local minima and improving optimization. However, this conclusion was drawn only using simple OOD data types like rotation.

Silva \etal~\cite{de2023value} and Geiping \etal~\cite{geiping2022much} show that, for small domain shifts, adding OOD data reduces generalization error on the original test set and exhibits non-monotonic behavior. While most research has relied on basic models (e.g., ResNet~\cite{resnet}) and datasets (e.g., MNIST~\cite{mnist}), our work seeks to produce OOD data study to more complex scenarios with diffusion models, utilizing advanced classification architectures to deepen the understanding of OOD effects.

\begin{figure*}[ht!]
\begin{center}
\includegraphics[width=1\linewidth]{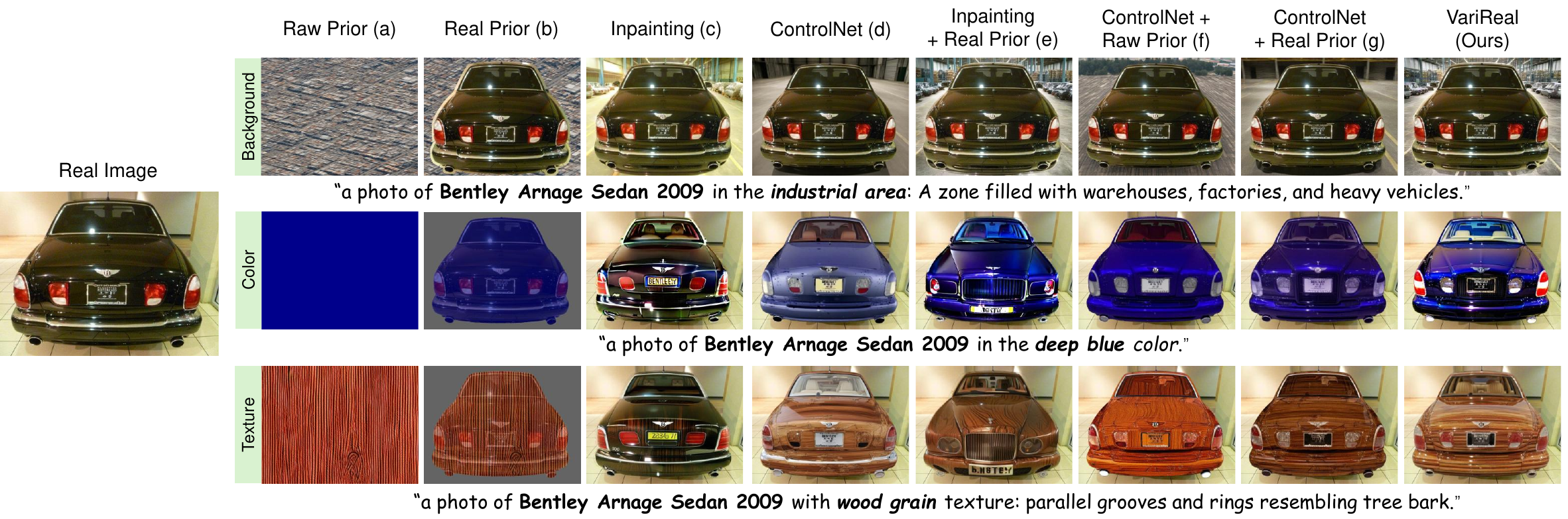}
\end{center}
\vspace{-1.8em}
   \caption{We compare images generated by various candidate methods: Inpainting model~\cite{sdinpaint} alone, ControlNet~\cite{controlnet} alone, Inpainting model with Real Prior, ControlNet with Raw Prior or Real Prior, and our final results for three attribute modifications. The first two columns illustrate the priors used (Raw Prior and Real Prior), and generation prompts used are listed beneath each image.}
\label{fig: method_compare}
\vspace{-1em}
\end{figure*}

\topic{Learning with synthetic data.} Several studies~\cite{kim2024datadream, he2022synthetic,yu2023diversify, da2023diversified,sariyildiz2023fake} focus on generating synthetic data that approximates real-world distributions. These approaches aim to create a dataset larger than the few-shot samples. Generated data supports various tasks, including object recognition~\cite{kim2024datadream, da2023diversified,dunlap2023diversify,sariyildiz2023fake}, object detection~\cite{feng2024instagen}, and semantic segmentation~\cite{wu2023diffumask}. Its effectiveness is demonstrated by training CLIP~\cite{clip} models exclusively on synthetic data or in combination with real data~\cite{scalinglaw, kim2024datadream, he2022synthetic}. As a result, we focus specifically on object classification using CLIP model.

\topic{Automatic approach for minimal change generation. } Unlike synthetic data generation methods that focus on creating novel and diverse in-distribution images \cite{kim2024datadream}, minimal change generation aims only to modify specific areas or attributes of existing real images. Generative models, particularly diffusion-based approaches \cite{dalle,imagen,glide,sd}, facilitate efficient image editing without requiring manual annotation \cite{he2022synthetic} or physical graphics engines \cite{cleardepth,infinigen}. In particular, text-to-image stable diffusion methods are popular for minimal-change editing due to their high fidelity generation. Beyond text guidance, these models also support diverse conditioning inputs, such as reference images through IP-Adaptor~\cite{ip-adapter} and Canny edge maps through ControlNet~\cite{controlnet}.

These methods fall into two main categories: fine-tuning needed approaches~\cite{instructpix2pix,hive,mgie}, and non-fine-tuning needed approaches such as attention- or mask-based diffusion methods~\cite{p2p,fpe}. Fine-tuned methods, such as InstructPix2Pix~\cite{instructpix2pix}, require model retraining to achieve desired edits across new input domains. In contrast, attention- and mask-based diffusion models can target specific modifications without further fine-tuning. Attention-based methods, like FPE~\cite{fpe} and P2P~\cite{p2p}, substitute certain self- or cross-attention layers in the U-Net~\cite{unet}’s denoising process, leveraging the interpretability of attention maps. However, these methods may not perform well in all scenarios, particularly with real images~\cite{fpe}. Mask-based diffusion models, such as inpainting methods~\cite{diffusioninpainting, contextdiffusion, paintbyexample, diffblender}, offer strong generalization and method versatility by enabling controlled edits within specified areas while preserving unmasked regions. However, when modifying objects itself, these models may occasionally alter subtle shape details. Methods like ControlNet~\cite{controlnet} can help maintain an object’s original structure during edits.

The most closely related work is VisMin~\cite{vismin}, which generates minimal-change data to improve vision-language model comprehension. However, VisMin does not support controlled edits across our targeted three attributes. In contrast, we introduce an automatic, off-the-shelf approach enabling minimal, photorealistic edits for arbitrary combinations of real images and textual instructions.

\section{Method}
\label{sec:method}

\begin{figure*}[t]
\begin{center}
\includegraphics[width=1\linewidth]{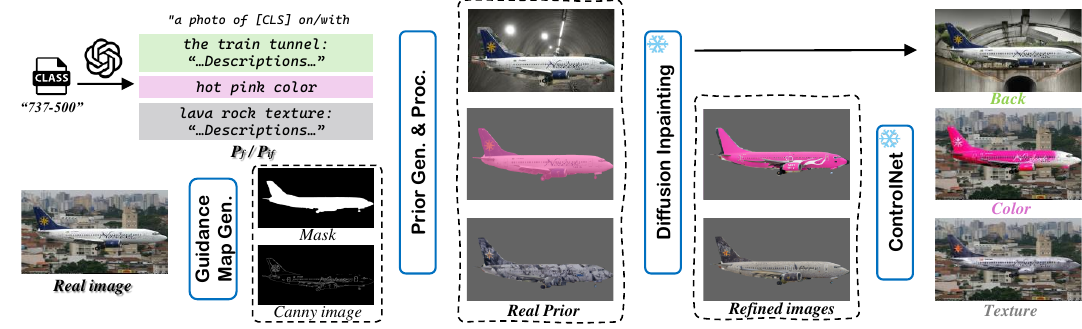}
\end{center}
\vspace{-1.8em}
   \caption{\textbf{An overview of VariReal pipeline.} Minimal-change steps for background, color, and texture are highlighted in green, pink, and grey, respectively. Real images are processed to generate guidance maps (e.g., masks, Canny edges) for Inpainting and ControlNet. GPT-4 generates feasible and infeasible prompts ($P_{f}$ and $P_{if}$), which guide color retrieval or prior image generation via Stable Diffusion. These Real Priors, combined with masks and prompts, are input to the inpainting model. For color and texture, ControlNet with Canny conditioning ensures precise foreground shapes. A final refinement step produces the optimal output for each setting.}
\label{fig: main_method}
\vspace{-1em}
\end{figure*}

\subsection{Preliminaries}

\topic{Task formulation.} Our goal is to analyze the impact of feasible and infeasible synthetic data ($I_{\text{Syn}}$), with feasibility defined per individual class $c_i$, where $i \in {1, \dots, C}$. Our VariReal method generates minimal-change $I_{\text{Syn}}$ pairs from a shared real-image base ($I_{\text{Real}}$) using distinct textual prompts. Our approach isolates feasibility across three targeted attribute categories—background, color, and texture—while minimally altering other image content (e.g., the same dog depicted with different colors). The textual guidances are class-specific, LLM-generated prompts categorized as feasible ($P_f$) and infeasible ($P_{if}$). Each real image ($I_{\text{Real}}$) is combined with all prompts from both categories, ensuring every real image is repeated equally, $|P_f| = |P_{if}|$. By varying the number of prompts ($|P_f| \geq 1$), we assess the impact of additional synthetic augmentations. Note that the texture attribute inherently includes color characteristics. Finally, we LoRA fine-tune CLIP models on in-distribution and OOD synthetic datasets to compare how each data type influences downstream classification performance.

\topic{Fine-tuning with low-rank adaptation.} The Low-Rank Adaptation~\cite{lora} introduces low-rank decomposition into the pre-trained weight matrix to reduce the number of learnable parameters. The final weights after fine-tuning could be expressed by $h = W_0x + BAx$, where $W_0$ represents the pre-trained weights. The decomposed weights $B \in \mathbb{R}^{d \times r}$ and $A \in \mathbb{R}^{r \times k}$, with LoRA rank $r \ll \min(d, k)$.

\topic{Latent diffusion models. } 
Latent Stable Diffusion~\cite{sd} encodes an image into a latent space using an encoder, defined as \( z_0 = E(x_0) \), and learns a conditional distribution \( p(z|c) \) by predicting the Gaussian noise added to the latent vector. The objective function can be expressed as:
\vspace{-1mm}
\begin{equation}
\min_{\theta} \, \mathbb{E}_{(x, c) \sim \mathcal{D}, \, \epsilon \sim \mathcal{N}(0,1), \, t} \left[ \left\| \epsilon - \epsilon_{\theta}(z_t, c, t) \right\|_2^2 \right]
\vspace{-1mm}
\end{equation}
where \( z_t \) is the noisy latent representation, $c$ is corresponding conditions and \( \epsilon \) represents the Gaussian noise added at each time step \( t \). For the inference process, a randomly noised vector is sampled and denoised over total \( T \) steps to obtain the final latent representation \( z_0 \), which is then decoded back into pixel space using the decoder \( D(z_0) \) of the VAE~\cite{vae}. 

\topic{A naive solution.}Naive solutions could employ text-guided Inpainting models~\cite{diffusioninpainting} (e.g., SDXL Inpainting) or Canny-edge-based ControlNet~\cite{controlnet} models (e.g., SDXL ControlNet), using a base prompt \( P_{\text{base}} = \text{"a photo of a [CLS]"} \). Inpainting methods generate more natural images but are heavily influenced by the original attributes, limiting their effectiveness for substantial changes, as illustrated by the persistent dark hue when changing a black car's color of Figure~\ref{fig: method_compare} (c). 

Conversely, ControlNet preserves object structure independently from original attributes in Figure~\ref{fig: method_compare} (d) but often produces less natural edits for color and texture and can cause objects to appear floating when modifying backgrounds, reducing realism.

\topic{Motivation.}To overcome the limitations of existing methods, we design a pipeline which overcomes the individual weaknesses of out-of-the-box methods by combining the individual strengths, i.e. combining Inpainting's realism with ControlNet's preciseness. 

\subsection{VariReal: Generating minimal-change data}
\label{chap:gen}
We present a zero-shot pipeline for minimal-change image generation. \cref{sec: promptgen} details prompt generation \( P_{f} \) and \( P_{if} \), followed by our prior-based generation process in \cref{sec: mainmethod}, including key steps like guidance maps and final processing. We also compare candidate models to determine the optimal modification strategy. Lastly, \cref{sec: filtering} covers MLLM-based filtering.

\subsubsection{Guidance prompt}
\label{sec: promptgen}
$P_{f}$ and 
$P_{if}$ are as text prompts for Stable Diffusion model to guide desired content. 
To generate as many accurate $P_{f}$ and $P_{if}$ per fine-grained class as possible, we utilize ChatGPT-4~\cite{gpt4} with In-Context Learning~\cite{incontext}, providing the model with positive examples $Example+$ and negative examples $Example-$ to help avoid errors and repetitive content. To improve the fine-grained detail and realism of the generated backgrounds or textures, we instruct GPT to append a brief explanatory description when generating prompts, providing more detailed guidance for image generation.

Although large language models possess broad knowledge across various domains, ChatGPT still regularly designates attributes as 'feasible' for a target object that do not exist in the real world, particularly for fine-grained classes for which it has limited knowledge. For example, fine-grained airplane class "737-500" normally do not have color in purple. To address this issue, we design additional prompts to instruct the model to perform preliminary checks and filtering on its outputs. Manual verification ensures that feasible prompts align with the training domain. Using the same base prompt and ChatGPT-generated results, we form our final prompts shown in the Figure~\ref{fig: teaser} and Figure~\ref{fig: method_compare}. Details of the generation process and filtered ratios are provided in the Supplementary \cref{sec:guidance_prompt}.

\subsubsection{Prior-guided minimal change generation}

\label{sec: mainmethod}
\phantomsection
\label{sec:groundingmask}
\topic{Guidance map generation.} The guided mask and canny images are for inpainting model and ControlNet respectively. We use Grounding DINO~\cite{groundingdino} to generate bounding boxes \( \text{bbox}_i \), which are then fed into the SAM2~\cite{sam2} model to produce masks \( m_i \) for each category \( c_i \). For samples without detectable bounding boxes, we use the RMBG1.4~\cite{rmbg} foreground segmentation model as a fallback to ensure each sample has a mask.

In our method, we use the Canny image ControlNet~\cite{controlnet} model. For all settings, the Canny image is created by extracting the foreground \( \text{Foreground}_i \) from \( \text{mask}_i \).

\phantomsection
\label{sec: prior}
\topic{Prior generation and process.} 
We use prompts \(P_{f}\) or \(P_{if}\) with Stable Diffusion to generate “Background” and “Texture” Priors, and predefined RGB values from a Color Bank for the “Color Prior”. These initial outputs are termed \textit{Raw Priors}.

To integrate these priors with real images, we merge the original object's region with the Background Prior, applying mask dilation to preserve spatial context and realism (e.g., ensuring pets remain grounded). We ablate this operation effect in the Supplementary \cref{sec: app_ablation}. For color and texture edits, the generated prior is overlaid via an alpha channel to retain the original shape and details of the subject. These refined results are referred to as \textit{Real Priors}. The Figure~\ref{fig: method_compare} (a-b) illustrate these Priors. ControlNet leverages both Raw and Real Priors as conditions via IP-Adaptor~\cite{ip-adapter}, whereas Inpainting exclusively employs Real Prior to retain unchanged original information.

\phantomsection
\label{sec: finalprocess}
\topic{Final process.}
Before outputting the final images, the last step involves copying invariant regions from the original image and pasting them onto the generated image, ensuring minimal alterations.

\topic{Minimal change for background.} 
Figure~\ref{fig: method_compare} (e) demonstrates that incorporating prior information significantly enhances background editing quality, fulfilling our minimal-change requirement. Our optimal results are obtained using Inpainting with the Real Prior, a background-region mask, and the corresponding prompt \(P\) shown in Figure~\ref{fig: main_method}.

\phantomsection
\label{sec: detailedmethod}
\topic{Minimal change for foreground.} 
In contrast, color and texture edits require foreground modifications.  As shown in Figure~\ref{fig: method_compare} (e-g), single-stage Inpainting and ControlNet models are insufficient under either Raw or Real Priors: Inpainting may distort object shapes, while ControlNet can produce unnatural results. To address this, we first produce an initial refined image using SDXL Inpainting, then use it as a conditional input for ControlNet to generate the final image. This combined approach (Figure~\ref{fig: main_method}) leverages the strengths of both methods, preserving the object's shape while achieving natural and precise color or texture changes.

\subsubsection{Automatic filtering}
\label{sec: filtering}
To ensure generated images meet prompt requirements, the MLLM Llava-Next~\cite{llavanext} model checks each image's feasibility and attributes. Using predefined questions, we filter out images that do not match the specified background, color, or texture. More details and example about the filtering questions can be found in the Supplementary \cref{sec: app_af}.

\subsection{Feasibility effectiveness validation}
\label{chap:clip}

Following the common practice~\cite{scalinglaw, kim2024datadream} to evaluate the impact of data feasibility, we fine-tune a CLIP~\cite{clip} classifier, which encodes images and corresponding text prompts to calculate similarity scores for classification. We apply LoRA~\cite{lora} modules to fine-tune both CLIP's image and text encoders. For each class \( c_i \in C \), we use the prompt "a photo of [CLS]" as text input. Training is performed via supervised learning using cross-entropy loss, updating only the LoRA modules while keeping pretrained weights frozen.

In mixed training scenarios (real and synthetic data), the loss is a weighted combination defined as:
\vspace{-2mm}
\begin{equation}
\mathcal{L}_{C} = \lambda \cdot \text{CE}(\text{Real}) + (1 - \lambda) \cdot \text{CE}(\text{Synth})
\vspace{-2mm}
\end{equation}

where \( \lambda \) balances the contribution from real data, and \(\text{CE}\) denotes cross-entropy loss.

\section{Experiments}
\label{sec:experiments}

\begin{table*}[t]
\centering
\setlength{\tabcolsep}{1.5pt}
\resizebox{\textwidth}{!}{
\renewcommand{\arraystretch}{1.3}
\begin{tabular}{@{\extracolsep{10pt}}lllcccccccccccccccccccc@{}}
\toprule
& \textbf{R} & \textbf{S} &
\multicolumn{5}{c}{\textbf{Pets~\cite{pets}}} &
\multicolumn{5}{c}{\textbf{AirC~\cite{aircraft}}} &
\multicolumn{5}{c}{\textbf{Cars~\cite{cars}}} &
\multicolumn{5}{c}{\textbf{Average}} \\ 
 & & & F & IF & Mix & $\Delta_1$ & $\Delta_2$ 
         & F & IF & Mix & $\Delta_1$ & $\Delta_2$
         & F & IF & Mix & $\Delta_1$ & $\Delta_2$
         & F & IF & Mix & $\Delta_1$ & $\Delta_2$ \\
\cline{1-3} \cline{4-8} \cline{9-13} \cline{14-18} \cline{19-23}
\textit{0-shot} & & &
\multicolumn{5}{c}{--- 91.0 ---} & 
\multicolumn{5}{c}{--- 23.8 ---} & 
\multicolumn{5}{c}{--- 63.2 ---} & 
\multicolumn{5}{c}{--- 59.3 ---} \\ 
\textit{Real} & \ding{51} & &
\multicolumn{5}{c}{--- 95.2 ---} & 
\multicolumn{5}{c}{--- 84.5 ---} & 
\multicolumn{5}{c}{--- 92.6 ---} &  
\multicolumn{5}{c}{--- 90.8 ---} \\
\cline{1-3} \cline{4-8} \cline{9-13} \cline{14-18} \cline{19-23}
Back. & & \ding{51} &
95.4 & 95.3 & 95.2 & \posval{+0.1} & \negval{-0.2} &
86.8 & 85.0 & 87.1 & \posval{+1.8} & \posval{+1.2} &
93.7 & 93.8 & 93.8 & \textcolor{red}{-0.1} & \posval{+0.1} &
92.0 & 91.4 & 92.0 & \posval{+0.6} & \posval{+0.4} \\
Color & & \ding{51} &
94.5 & 94.4 & 94.1 & \posval{+0.1} & \negval{-0.4} & 
80.8 & 81.6 & 81.9 & \negval{-0.8} & \posval{+0.7} &
91.6 & 91.5 & 91.6 & \posval{+0.1} &\posval{+0.1} &
89.0 & 89.1 & 89.2 & \negval{-0.1} & \posval{+0.2} \\
Text. &  & \ding{51} &
93.8 & 93.3 & 92.8 & \posval{+0.5} & \negval{-0.8} &
81.6 & 81.9 & 82.0 & \negval{-0.3} & \posval{+0.3} &
90.9 & 87.7 & 91.8 & \posval{+3.2} & \posval{+3.0} &
88.8 & 87.6 & 88.9 & \posval{+0.2} & \posval{+0.7} \\ 
\cline{1-3} \cline{4-8} \cline{9-13} \cline{14-18} \cline{19-23}
Back. & \ding{51} & \ding{51} &
95.3 & 95.3 & 95.3 & \posval{+0.0} & \posval{+0.0} &
88.0 & 88.4 & 88.6 & \negval{-0.4} & \posval{+0.4} &
93.8 & 93.7 & 93.6 & \posval{+0.1} & \negval{-0.2} &
92.4 & 92.5 & 92.5 & \negval{-0.1} & \posval{+0.1}  \\
Color & \ding{51} & \ding{51} &
95.3 & 95.2 & 95.0 & \posval{+0.1} & \negval{-0.3} &
84.6 & 84.0 & 83.6 & \posval{+0.6} & \negval{-0.7} &
92.7 & 92.5 & 92.8 & \posval{+0.2} & \posval{+0.2} &
90.9 & 90.5 & 90.4 & \posval{+0.4} & \negval{-0.2} \\
Text. & \ding{51} & \ding{51} &
95.3 & 95.2 & 95.2 & \posval{+0.1} & \negval{-0.1} &
83.9 & 83.8 & 83.8 & \posval{+0.1} & \negval{-0.1} &
93.0 & 92.8 & 92.6 & \posval{+0.2} & \negval{-0.3} &
90.7 & 90.6 & 90.5 & \posval{+0.1} & \negval{-0.1} \\ 
\bottomrule
\end{tabular}
}
\vspace{-0.5em}
\caption{Top-1 performance using the full training set and synthetic images generated by VariReal, including baseline, synthetic-only and synth + real. The number of synthetic images is set to five times the number of real images across all experiments. R/S indicates real/synthetic fine-tuning. F/IF denotes feasible/infeasible inputs, Mix indicates training with both. $\Delta_1 = F - IF$, and $\Delta_2 = Mix - \frac{F + IF}{2}$ measures the gain/loss of mixing compared to the average of individual setting.}
\label{tab: main1}
\vspace{-1.0em}
\end{table*}

\subsection{Experiments setup}
\label{sec:dataset}
\topic{Dataset.}Our synthetic data for background, color, and texture modifications require images with clearly defined foreground objects and visible backgrounds; hence, datasets dominated by foreground-only images, such as ImageNet~\cite{imagenet}, are unsuitable. Fine-grained datasets offer clearer comparisons between feasible and infeasible attribute variations. Therefore, we generate our minimal-change synthetic datasets from three fine-grained sources: Oxford Pets~\cite{pets}, FGVC Aircraft~\cite{aircraft}, and Stanford Cars~\cite{cars}. Additionally, to specifically evaluate background modifications, we use the binary classification WaterBirds dataset~\cite{dunlap2023diversify}, which pairs landbirds and waterbirds with water or land backgrounds.

\topic{Implementation details.}
Our VariReal pipeline utilizes SDXL Inpainting v0.1 and SDXL ControlNet v1.0~\cite{controlnet} based on Canny-edge conditioning, along with Stable Diffusion v2.1~\cite{sd} for prior image generation in background and texture modifications. The Llava-1.6-7B~\cite{llava} model is employed for automatic filtering. Real images used for modification are sourced from the training split of each dataset, and performance is evaluated on the original test set. We use $|P_{f}| = |P_{if}| = 5$ prompts per class, thus generating five synthetic images per real-image base.

We fine-tune a CLIP ViT-B/16~\cite{vit} classifier using LoRA modules with the rank of 16 applied to both image and text encoders, optimized with AdamW~\cite{adam}. The scale factor $\lambda$ is set to $0.5$ to equally weight real and synthetic cross-entropy losses. To ensure fair training budget despite varying dataset sizes (real-only, synthetic-only, and mixed synth+real training), we fix the total maximum training iterations to ensure same optimizer update steps. Detailed generation and training hyperparameters are provided in the Supplementary \cref{sec: app_details}.

\topic{Baseline methods.}
To evaluate the impact of feasible versus infeasible synthetic data, we use zero-shot CLIP and CLIP fine-tuned on real images as baselines. We compare these baselines with CLIP trained exclusively on synthetic data and on combinations of synthetic and real data.

\topic{Evaluation protocol.}
We measure classification performance using top-1 accuracy (\%). For dataset distribution analysis in \cref{sec:distribution_any}, we report FID~\cite{fid}, CLIP score~\cite{clip}, DINO score~\cite{dinov2}, and LPIPS~\cite{lpips} scores. More details on those metrics are described in \cref{sec:distribution_any}.

\subsection{Classification with minimally changed data}

\subsubsection{The role of feasibility}
\label{sec: ex_train_all_data}

Table~\ref{tab: main1} compares model performance across four training settings: (1) two baselines, (2) synthetic-only, and (3) real + synthetic training. To assess the role of feasibility, we define the metric $\Delta_1 = F - IF$, where $F$ and $IF$ denote performance using feasible and infeasible data, respectively. As shown in the second-to-last column, 4 out of 6 cases yield positive $\Delta_1$, suggesting that feasible data generally performs slightly better than infeasible.

Under the synthetic-only setting, 56\% of $\Delta_1$ values (5/9) fall within 0.3\% across each dataset column. Specifically, in the AirC~\cite{aircraft} dataset, feasible data outperforms infeasible by 1.8\% under the background setting, while infeasible data performs better by 0.8\% and 0.3\% under the color and texture settings, respectively. After incorporating real data (real + synthetic), 78\% of $\Delta_1$ values (5/9) remain within 0.3\%, indicating that the performance gap between feasible and infeasible data is consistently small across settings.

\begin{observationbox}
\textit{Observation 1: Although feasible images perform slightly better, feasibility shows no clear impact on classification performance.}
\end{observationbox}

\subsubsection{The role of attribute}
\label{sec:ex_main}
Although all settings in Table~\ref{tab: main1} outperform the zero-shot baseline, synthetic color and texture data remain less effective compared to the real data. For instance, in the synthetic-only setting, feasible and infeasible color data achieve 89.0\% and 89.1\% average performance, both below the real-only fine-tuning baseline of 90.8\%. Even when combined with real data, color edits perform slightly worse by 0.1\% and 0.2\%.

In contrast, background modifications consistently improve performance. For instance, under synthetic-only training, feasible and infeasible backgrounds yield average accuracy gains of 1.2\% and 0.6\%, respectively, and 1.6\% and 1.7\% in the real + synthetic setting.

We further validate the benefits of background modifications on the WaterBirds~\cite{dunlap2023diversify} dataset (see Supplementary \cref{sec: app_waterbird}). Both feasible and infeasible background edits outperform real-only setting, with improvements of 0.9\% and 6.7\% respectively in the synthetic-only setting, and 7.2\% and 8.8\% in the real + synthetic setting.

\begin{observationbox}
\textit{Observation 2: Compared to fine-tuning on real data alone, adding synthetic data with background modifications improves performance, whereas synthetic foreground edits (color and texture) are less effective.}
\end{observationbox}

\subsubsection{The role of mixed training}

To assess the effect of mixing feasible and infeasible data, we construct a balanced synthetic dataset (third column of each subtable in Table~\ref{tab: main1}), with a total size five times that of the real training set. We define the metric $\Delta_2 = \text{Mix} - \frac{F + IF}{2}$ to measure the performance gain from mixing, relative to the average of using feasible and infeasible data separately.

In the real + synthetic setting, mixed training yields comparable performance, with average $\Delta_2$ deviations within 0.2\%. In contrast, under the synthetic-only setting, mixing leads to greater gains—0.4\%, 0.2\%, and 0.7\% improvements on average for background, color, and texture edits—indicating stronger complementarity between feasible and infeasible data. Further analysis for this is provided in Supplementary \cref{sec: app_classanyalysis}. This suggests that, unlike ALIA~\cite{dunlap2023diversify}, modifications need not be strictly feasible.

\begin{observationbox}
\textit{Observation 3: It is not necessary to strictly generate only feasible synthetic images to achieve performance gain.}
\end{observationbox}

\subsection{Analysis of minimally changed data}
\subsubsection{Qualitative results}

\begin{figure}[t]
\centering
\includegraphics[width=1\linewidth]{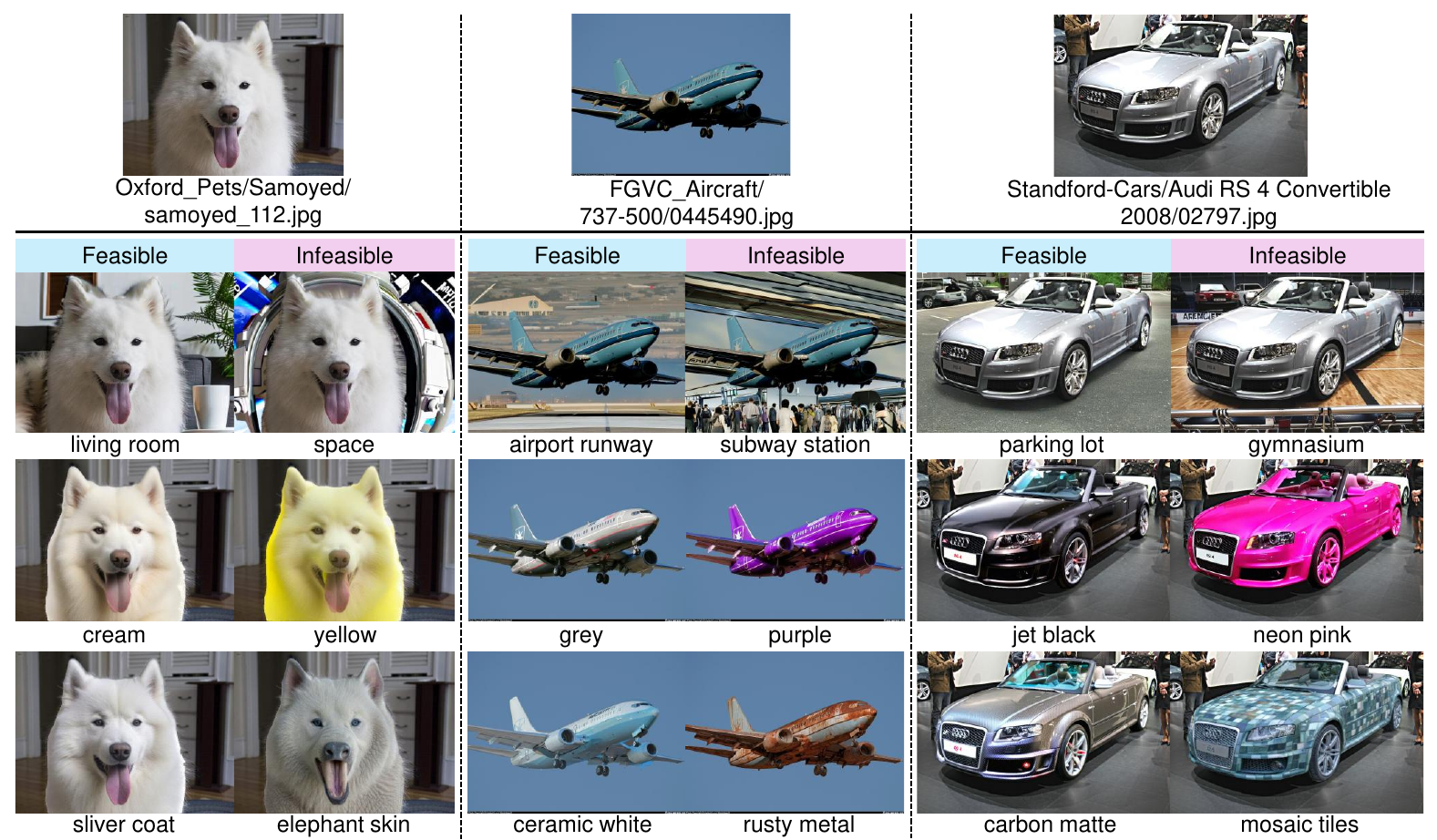}
\vspace{-1.8em}
\caption{Selected generation results from the three datasets. Only target prompt keywords are shown; detailed background and texture descriptions are omitted. Please zoom in for visual details.}
\vspace{-1em}
\label{sample_1}
\end{figure}

To assess the quality of VariReal-generated images, Figure~\ref{sample_1} presents qualitative examples from all three datasets. These examples demonstrate that the edits follow the text prompts with minimal changes and align with our feasibility definition—existing in real world. For instance, “neon pink” is not a released color for the “Audi RS 4 Convertible 2008” and is thus treated as infeasible. The images show the expected modification, while the rest of the image remains unchanged from the real source. More examples are provided in Supplementary \cref{sec: app_quaex}.

To further validate image quality, we conducted a user study with six human annotators using a questionnaire. Evaluators assessed each image on two aspects: (1) feasibility—whether feasible images appear realistic and infeasible ones do not—and (2) naturalness, rated on a 1–5 scale, where 5 indicates the most natural appearance. More details about the scoring setup are included in Supplementary \cref{sec: app_quaex}.

Feasibility is central to our pipeline, ensuring a clear distinction between feasible and infeasible subsets. As shown in Table~\ref{tab: human_study}, feasibility correctness is high, with error rates below 8\% for feasible and 16\% for infeasible data. The slightly lower accuracy for infeasible cases stems from occasional mismatched background-object combinations and difficulty capturing fine-grained texture details—particularly in the AirC dataset, as noted in annotator feedback (see Supplementary \cref{sec: app_quaex}). These results support the effectiveness of our approach, with VariReal reliably generating high-quality edits, further refined by automatic filtering (\cref{sec: filtering}).

Regarding how natural the generated images are, VariReal images received acceptable naturalness scores from human annotators—averaging 3.94 for feasible and 3.96 for infeasible data. For failure cases, some generated images appear less natural (see Supplementary \cref{sec: app_quaex}) because of a dramatic change from the original color to a new color, such as red to white.

\begin{table}[h]
\center
\renewcommand{\arraystretch}{0.8} 
\small 
\resizebox{0.9\columnwidth}{!}{%
\begin{tabular}{c|cccccc|cc}
\hline
\multirow{2}{*}{} &
  \multicolumn{2}{c}{\multirow{2}{*}{Back}} &
  \multicolumn{2}{c}{\multirow{2}{*}{Color}} &
  \multicolumn{2}{c|}{\multirow{2}{*}{Texture}} &
  \multicolumn{2}{c}{\multirow{2}{*}{Averaged}} \\
                      & \multicolumn{2}{c}{} & \multicolumn{2}{c}{} & \multicolumn{2}{c|}{} & \multicolumn{2}{c}{} \\
\multicolumn{1}{l|}{} & F        & IF        & F        & IF        & F         & IF        & F        & IF        \\ \hline
\multirow{2}{*}{\begin{tabular}[c]{@{}c@{}}Feasibility \\ Correctness/\%\end{tabular}} &
  \multirow{2}{*}{92.1} &
  \multirow{2}{*}{87.5} &
  \multirow{2}{*}{94.4} &
  \multirow{2}{*}{85.2} &
  \multirow{2}{*}{90.1} &
  \multirow{2}{*}{80.9} &
  \multirow{2}{*}{92.2} &
  \multirow{2}{*}{84.2} \\
                      &          &           &          &           &           &           &          &           \\ \hline
\multirow{2}{*}{\begin{tabular}[c]{@{}c@{}}Naturalness\\ Score(0.0-5.0)\end{tabular}} &
  \multirow{2}{*}{4.5} &
  \multirow{2}{*}{4.1} &
  \multirow{2}{*}{3.62} &
  \multirow{2}{*}{3.90} &
  \multirow{2}{*}{3.70} &
  \multirow{2}{*}{3.88} &
  \multirow{2}{*}{3.94} &
  \multirow{2}{*}{3.96} \\
                      &          &           &          &           &           &           &          &           \\ \hline
\end{tabular}%
}

\vspace{-0.5em}
\caption{Human evaluation of the generated dataset based on feasibility correctness and naturalness scores, validating its suitability for downstream tasks.}
\label{tab: human_study}
\vspace{-1.0em}

\end{table}

\subsubsection{Distribution analysis}
\label{sec:distribution_any}
We analyze the dataset using several similarity metrics to better understand the distributional differences between feasible and infeasible data and their relation to in- and out-of-distribution. We compute the Fréchet Inception Distance (FID)~\cite{fid} to quantify the distributional similarity between generated and real data. Additionally, we use: \textbf{CLIP Score}: calculated cosine similarity for feature from the ViT-L/14 model~\cite{vit}. \textbf{DINO Score}: computed cosine similarity for feature from the DINOv2-Base model~\cite{dinov2} for feature extraction. And \textbf{LPIPS Score}~\cite{lpips}.

\begin{figure}[t]
\centering
\includegraphics[width=1\linewidth]{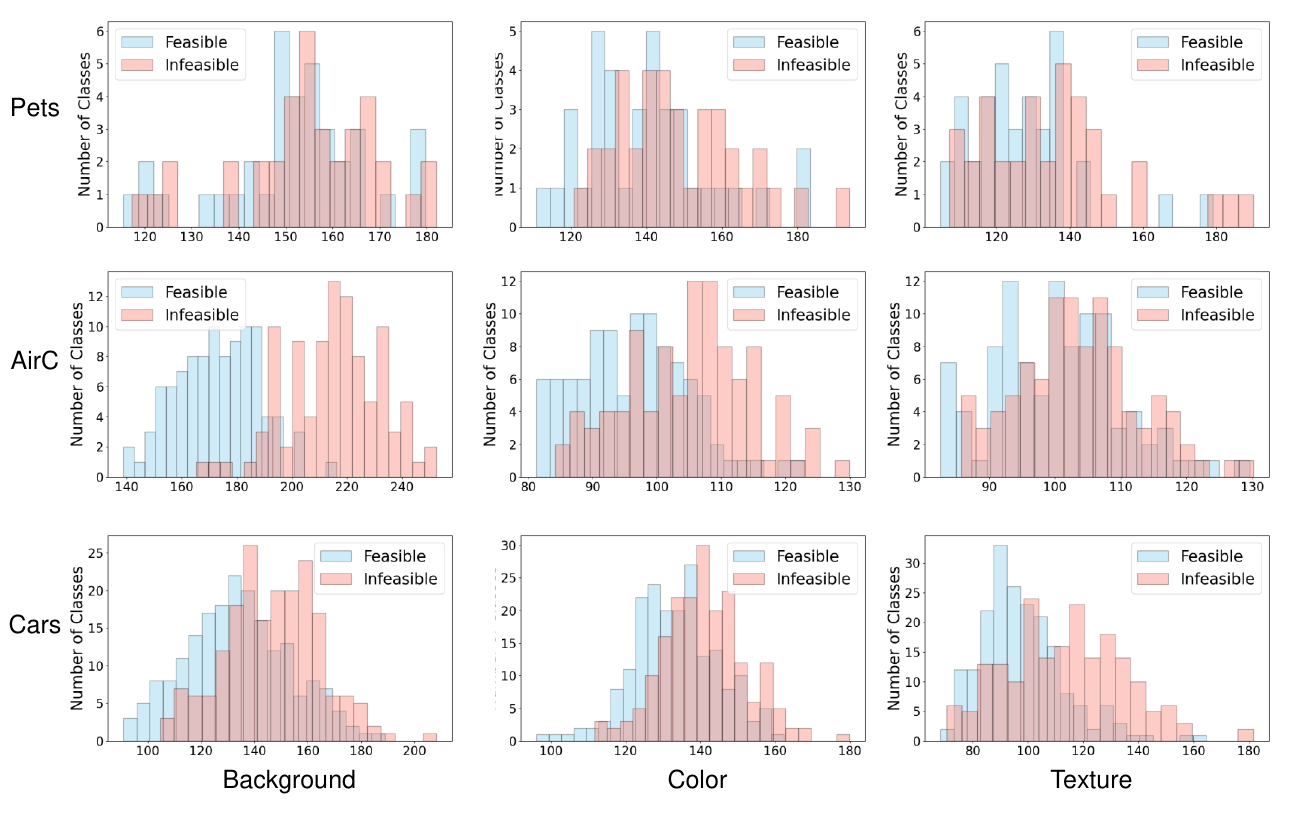}
\vspace{-1.8em}
\caption{The FID score settings compared using feasible and infeasible settings across different datasets.}
\label{fid}
\end{figure}

Figure~\ref{fid} shows that feasible samples generally resemble in-distribution data more closely than infeasible ones, aligning better with the real data distribution. This observation is supported by the metrics in Table~\ref{tab: main4}, which reports average scores across the three datasets. All three metrics indicate that feasible data is closer to real data. While CLIP and DINO scores show limited sensitivity to fine-grained differences, LPIPS captures subtle variations more effectively.

Interestingly, both feasible and infeasible foreground modifications (color and texture) are closer to real data than background edits. For instance, in the AirC~\cite{aircraft} dataset, FID peak scores for foreground edits are much lower (around 95 and 110 for feasible/infeasible) than for background edits (around 170 and 220). Table~\ref{tab: main4} shows similar trends—for instance, the average DINO score for color is about 10\% higher than for background. However, as discussed in \cref{sec:ex_main}, only background modifications consistently improve classification performance. This highlights the following:

\begin{observationbox}
\textit{Observation 4: Classification tasks are object-centric: although foreground (color and texture) modifications align more closely with real data distributions, changing them may deviate from meaningful class-relevant features, leading to weaker effects.}
\end{observationbox}

\begin{table}[h]
\center
\renewcommand{\arraystretch}{0.8} 
\small 
\resizebox{0.9\columnwidth}{!}{%
\begin{tabular}{c|c|ccc}
\toprule
\textbf{Settings}                      & \textbf{Inputs} & \textbf{CLIP} (↑) & \textbf{DINO} (↑) & \textbf{LPIPS} (↓) \\ \midrule
                             & F & \textbf{0.914}       & \textbf{0.861}       & \textbf{0.447}  \\
\multirow{-2}{*}{Background} & IF  & 0.886       & 0.830       & 0.477  \\ \midrule
                             & F & \textbf{0.951}       & \textbf{0.956}       & \textbf{0.189}  \\
\multirow{-2}{*}{Color}      &  IF & 0.904       & 0.939       & 0.254  \\ \midrule
                             & F &\textbf{ 0.936}       & \textbf{0.949}       & \textbf{0.207}  \\
\multirow{-2}{*}{Texture}    & IF & 0.898       & 0.925      & 0.218  \\ \bottomrule
\end{tabular}%
}
\caption{The average DINO, CLIP and LPIPS scores calculated between generated synthetic image and corresponding real images for three datasets. F/IF denotes feasible/infeasible inputs.}
\label{tab: main4}
\end{table}

\subsubsection{Scaling the number of training images}  
To further understand the impact of synthetic data by VariReal, we conducted a scaling analysis on the AirC dataset~\cite{aircraft}, adjusting feasible/infeasible synthetic-to-real ratios from 1:1 to 5:1.

\begin{figure}[t]
\centering
\includegraphics[width=1.0\linewidth]{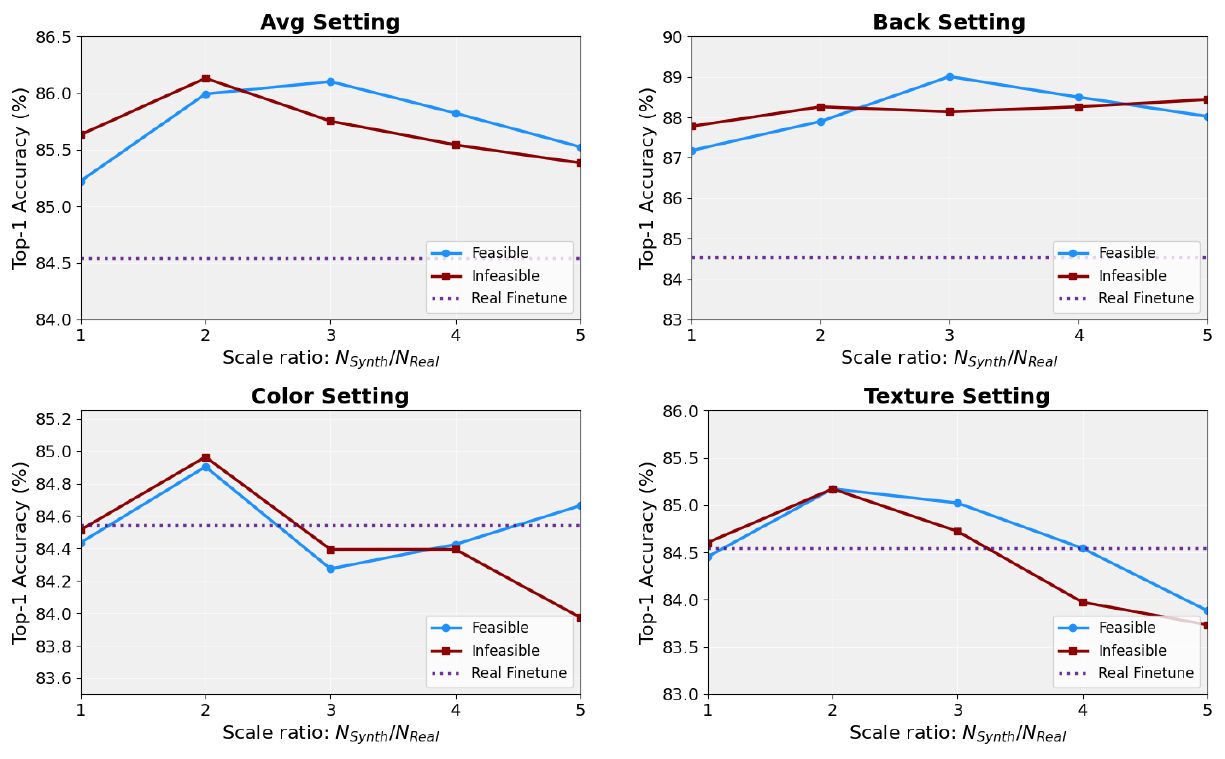}
\vspace{-1.5em}
\caption{The scaling experiment results for the FGVC-Aircraft~\cite{aircraft} dataset are shown for background, color, and texture settings. The horizontal axis represents the scale factor for synthetic images relative to real images. Here, the total real image training set is used, with scale factors ranging from 1 to 5.}
\label{scale_number}
\end{figure}

Our results in Figure~\ref{scale_number} reveal a nonlinear relationship between performance and data scale. While background modifications always benefit the downstream tasks, color and texture modifications achieve peak accuracy at smaller scales. Notably, performance slightly exceeds the baseline at these peaks but declines as more synthetic images are added. This indicates that both feasible and infeasible color and texture data behave similar to OOD data, while feasible data being relatively closer to the real distribution. Large-scale use of such data does not provide meaningful in-distribution information for downstream tasks. However, a limited amount can serve as effective augmentation, enhancing model performance and robustness.

\begin{observationbox}
\textit{Observation 5: Synthetic data with color and texture modifications can enhance classification performance as augmentation, but their effectiveness is limited to specific scaling ranges. In contrast, background modifications consistently yield performance gains.}

\end{observationbox}

\section{Conclusion}
\label{sec:conclusion}

In this work, we present VariReal, a pipeline for systematically investigating the impact of minimal-change feasible and infeasible synthetic data. By introducing controlled variations in background, color, and texture across three fine-grained datasets, we assess the role of feasibility through LoRA-based fine-tuning of a CLIP classifier. Our findings reveal a counter-intuitive result: feasibility does not significantly affect classification performance. Although typically assumed to benefit downstream tasks, feasible synthetic variations in color and texture are no more effective than real data—and in some cases, even degrade performance. In contrast, background modifications consistently improve accuracy, regardless of feasibility. This suggests that, for object-centric classification, altering foreground attributes may disrupt class-relevant signals and yield limited gains. Overall, our results underscore the nuanced effects of different attribute modifications and offer new insights for designing effective synthetic data generation strategies.

{
\noindent
\textbf{Acknowledgements.} Jae Myung Kim thanks the International Max Planck Research School for Intelligent Systems (IMPRS-IS)
and the European Laboratory for Learning and Intelligent Systems (ELLIS) PhD programs for support.
}



{
    \small
    \bibliographystyle{ieeenat_fullname}
    \bibliography{main}
}

\clearpage
\setcounter{page}{1}
\setcounter{section}{0}
\renewcommand{\thesection}{\Alph{section}}
\maketitlesupplementary

In this supplementary material, we first discuss the broader impacts and limitations of our analysis in \cref{sec:limitation}. Experimental setups for our method are provided in \cref{sec: app_details}, and configurations for other image editing models are detailed in \cref{app_others_detail}. \cref{sec: our_details} describes our method in detail, including guidance prompts and automatic filtering. Additionally, we present background-specific classification results on the WaterBird~\cite{dunlap2023diversify} dataset in \cref{sec: app_waterbird}. Further classification result analysis is provided in \cref{sec: app_classanyalysis}, followed by additional qualitative examples and user study details in \cref{sec: app_quaex}. Finally, an ablation study of the VariReal pipeline is included in \cref{sec: app_ablation}.

\section{Broader Impact and Limitation}
\label{sec:limitation}
\begin{table*}[t]
\centering
\resizebox{1.0\textwidth}{!}{%
\begin{tabular}{@{}c|cccccc|cccccc|cccccc}
\toprule
 &
  \multicolumn{6}{c|}{\textbf{Background}} &
  \multicolumn{6}{c|}{\textbf{Color(Per CLS)}} &
  \multicolumn{6}{c}{\textbf{Texture}} \\ \midrule
\rowcolor[HTML]{E7E7E7} 
 &
  \multicolumn{2}{c}{\cellcolor[HTML]{E7E7E7}Pets} &
  \multicolumn{2}{c}{\cellcolor[HTML]{E7E7E7}AirC} &
  \multicolumn{2}{c|}{\cellcolor[HTML]{E7E7E7}Cars} &
  \multicolumn{2}{c}{\cellcolor[HTML]{E7E7E7}Pets} &
  \multicolumn{2}{c}{\cellcolor[HTML]{E7E7E7}AirC} &
  \multicolumn{2}{c|}{\cellcolor[HTML]{E7E7E7}Cars} &
  \multicolumn{2}{c}{\cellcolor[HTML]{E7E7E7}Pets(Per CLS)} &
  \multicolumn{2}{c}{\cellcolor[HTML]{E7E7E7}AirC} &
  \multicolumn{2}{c}{\cellcolor[HTML]{E7E7E7}Cars} \\
 &
  F &
  IF &
  F &
  IF &
  F &
  IF &
  F &
  IF &
  F &
  IF &
  F &
  IF &
  F &
  IF &
  F &
  IF &
  F &
  IF \\ \midrule
\rowcolor[HTML]{FFFFFF} 
Raw output &
  50 &
  70 &
  50 &
  70 &
  50 &
  70 &
  10 &
  10 &
  10 &
  10 &
  10 &
  10 &
  8 &
  50 &
  30 &
  50 &
  15 &
  70 \\
Auto-filtering &
  47 &
  64 &
  36 &
  68 &
  44 &
  67 &
  6$\sim$7 &
  8$\sim$9 &
  7$\sim$8 &
  8$\sim$9 &
  7$\sim$8 &
  8$\sim$10 &
  7 &
  42 &
  25 &
  46 &
  12 &
  64 \\
\rowcolor[HTML]{FFFFFF} 
Manual-filtering &
  43 &
  50 &
  22 &
  50 &
  31 &
  50 &
  5 &
  5 &
  5$\sim$8 &
  5$\sim$6 &
  5 &
  5 &
  5 &
  27 &
  24 &
  44 &
  7 &
  57 \\ \midrule
\rowcolor[HTML]{E7E7E7} 
Final Accept Rate &
  0.86 &
  0.714286 &
  0.44 &
  0.71429 &
  0.62 &
  0.71429 &
  0.5 &
  0.5 &
  0.5$\sim$0.8 &
  0.5$\sim$0.8 &
  0.5 &
  0.5 &
  0.625 &
  0.54 &
  0.8 &
  0.88 &
  0.467 &
  0.814 \\ \bottomrule
\end{tabular}%
}
\caption{The number of prompts which are generated initially by LLM, after self-filtering and manual-filtering for each specific settings and some datasets. The Pets, AirC, Cars refer to our experimental dataset introduced in \ref{sec:dataset}.}
\label{tab: app1}
\end{table*}
Our VariReal pipeline focuses on generating feasible and infeasible image pairs for downstream tasks, with potential applications beyond classification. It offers a robust method for modifying backgrounds, colors, and textures in both prompts and real images, making it suitable for image editing tasks that require precise changes while preserving other regions. VariReal can also serve as a dataset generation tool to fine-tune Stable Diffusion models for text-guided image editing, enabling targeted modifications. Additionally, it supports data augmentation, showing that augmenting both feasible and infeasible backgrounds improves classification performance—unlike ALIA~\cite{dunlap2023diversify}, which only uses feasible backgrounds.

We define feasibility as alignment with real-world plausibility. For instance, feasible car colors are those officially released by manufacturers. Rare custom paint jobs—such as a "cyan" Audi RS 4 Convertible 2008—are excluded, as they do not reflect typical production offerings. Within our scope, such extreme cases are treated as infeasible settings.

Our approach targets datasets with clear foreground-background separation and focuses on classification tasks under minimal-change settings. Although we strive to preserve structure, slight deviations—particularly in color and texture edits—are sometimes unavoidable due to current image editing limitations. In the meantime, our method requires adjusting hyperparameters when modifying images to meet specific requirements. We believe advances in image editing techniques will make our experimental setup more effective and easier to implement. Due to resource constraints, we explored only three attributes (background, color, texture), but future work could extend to others, such as lighting. Developing a unified method for minimal, single-step edits across multiple attributes would enhance scalability and enable broader application to diverse datasets and tasks.

\section{Implementation Details}
\label{sec: app_details}
\begin{table*}[!htbp]
\centering
\resizebox{1.0\textwidth}{!}{%
\begin{tabular}{cc|cccccc|cccccc|cccccc}
\toprule
 &
   &
  \multicolumn{6}{c|}{Back.} &
  \multicolumn{6}{c|}{Color} &
  \multicolumn{6}{c}{Texture} \\ \midrule
\multicolumn{2}{c|}{\multirow{2}{*}{Parameters}} &
  \multicolumn{2}{c}{Pets} &
  \multicolumn{2}{c}{AirC} &
  \multicolumn{2}{c|}{Cars} &
  \multicolumn{2}{c}{Pets} &
  \multicolumn{2}{c}{AirC} &
  \multicolumn{2}{c|}{Cars} &
  \multicolumn{2}{c}{Pets} &
  \multicolumn{2}{c}{AirC} &
  \multicolumn{2}{c}{Cars} \\
\multicolumn{2}{c|}{} &
  F &
  IF &
  F &
  IF &
  F &
  IF &
  F &
  IF &
  F &
  IF &
  F &
  IF &
  F &
  IF &
  F &
  IF &
  F &
  IF \\ \midrule
\multicolumn{2}{c|}{\multirow{2}{*}{Guidance Scale for SDXL Inpainting~\cite{sdinpaint}}} &
  \multicolumn{2}{c}{\multirow{2}{*}{40}} &
  \multicolumn{2}{c}{\multirow{2}{*}{7.5}} &
  \multicolumn{2}{c|}{\multirow{2}{*}{7.5}} &
  \multicolumn{2}{c}{\multirow{2}{*}{12}} &
  \multicolumn{2}{c}{\multirow{2}{*}{12}} &
  \multicolumn{2}{c|}{\multirow{2}{*}{30}} &
  \multicolumn{2}{c}{\multirow{2}{*}{12}} &
  \multicolumn{2}{c}{\multirow{2}{*}{8}} &
  \multicolumn{2}{c}{\multirow{2}{*}{30}} \\
\multicolumn{2}{c|}{} &
  \multicolumn{2}{c}{} &
  \multicolumn{2}{c}{} &
  \multicolumn{2}{c|}{} &
  \multicolumn{2}{c}{} &
  \multicolumn{2}{c}{} &
  \multicolumn{2}{c|}{} &
  \multicolumn{2}{c}{} &
  \multicolumn{2}{c}{} &
  \multicolumn{2}{c}{} \\
\multicolumn{2}{c|}{\multirow{2}{*}{Guidance Scale for ControNet~\cite{controlnet}}} &
  \multicolumn{6}{c|}{\multirow{2}{*}{-}} &
  \multicolumn{6}{c|}{\multirow{2}{*}{7.5}} &
  \multicolumn{6}{c}{\multirow{2}{*}{7.5}} \\
\multicolumn{2}{c|}{} &
  \multicolumn{6}{c|}{} &
  \multicolumn{6}{c|}{} &
  \multicolumn{6}{c}{} \\
\multicolumn{2}{c|}{\multirow{2}{*}{Strength   for SDXL}} &
  \multicolumn{2}{c}{\multirow{2}{*}{0.99}} &
  \multicolumn{2}{c}{\multirow{2}{*}{0.95}} &
  \multicolumn{2}{c|}{\multirow{2}{*}{0.9}} &
  \multicolumn{2}{c}{\multirow{2}{*}{0.3}} &
  \multicolumn{2}{c}{\multirow{2}{*}{0.8}} &
  \multicolumn{2}{c|}{\multirow{2}{*}{0.85}} &
  \multirow{2}{*}{0.3} &
  \multirow{2}{*}{0.3} &
  \multirow{2}{*}{0.65} &
  \multirow{2}{*}{0.3} &
  \multirow{2}{*}{0.65} &
  \multirow{2}{*}{0.3} \\
\multicolumn{2}{c|}{} &
  \multicolumn{2}{c}{} &
  \multicolumn{2}{c}{} &
  \multicolumn{2}{c|}{} &
  \multicolumn{2}{c}{} &
  \multicolumn{2}{c}{} &
  \multicolumn{2}{c|}{} &
   &
   &
   &
   &
   &
   \\
\multicolumn{2}{c|}{\multirow{2}{*}{IP-Adptor~\cite{ip-adapter} Strength}} &
  \multicolumn{6}{c|}{\multirow{2}{*}{-}} &
  \multicolumn{2}{c}{\multirow{2}{*}{0.7}} &
  \multicolumn{2}{c}{\multirow{2}{*}{0.4}} &
  \multicolumn{2}{c|}{\multirow{2}{*}{0.4}} &
  \multirow{2}{*}{0.2} &
  \multirow{2}{*}{0.5} &
  \multirow{2}{*}{0.65} &
  \multirow{2}{*}{0.4} &
  \multirow{2}{*}{0.65} &
  \multirow{2}{*}{0.4} \\
\multicolumn{2}{c|}{} &
  \multicolumn{6}{c|}{} &
  \multicolumn{2}{c}{} &
  \multicolumn{2}{c}{} &
  \multicolumn{2}{c|}{} &
   &
   &
   &
   &
   &
   \\
\multicolumn{2}{c|}{\multirow{2}{*}{Inference Step for SD}} &
  \multicolumn{6}{c|}{\multirow{2}{*}{20}} &
  \multicolumn{6}{c|}{\multirow{2}{*}{-}} &
  \multicolumn{6}{c}{\multirow{2}{*}{15}} \\
\multicolumn{2}{c|}{} &
  \multicolumn{6}{c|}{} &
  \multicolumn{6}{c|}{} &
  \multicolumn{6}{c}{} \\
\multicolumn{2}{c|}{\multirow{2}{*}{Inference Step for SDXL Inpainting}} &
  \multicolumn{6}{c|}{\multirow{2}{*}{30}} &
  \multicolumn{6}{c|}{\multirow{2}{*}{20}} &
  \multicolumn{6}{c}{\multirow{2}{*}{20}} \\
\multicolumn{2}{c|}{} &
  \multicolumn{6}{c|}{} &
  \multicolumn{6}{c|}{} &
  \multicolumn{6}{c}{} \\
\multicolumn{2}{c|}{\multirow{2}{*}{Inference   Step for ControlNet}} &
  \multicolumn{6}{c|}{\multirow{2}{*}{-}} &
  \multicolumn{6}{c|}{\multirow{2}{*}{30}} &
  \multicolumn{6}{c}{\multirow{2}{*}{30}} \\
\multicolumn{2}{c|}{} &
  \multicolumn{6}{c|}{} &
  \multicolumn{6}{c|}{} &
  \multicolumn{6}{c}{} \\
\multicolumn{2}{c|}{\multirow{2}{*}{Mask dilated factor/alpha factor}} &
  \multicolumn{2}{c}{\multirow{2}{*}{120}} &
  \multicolumn{2}{c}{\multirow{2}{*}{50}} &
  \multicolumn{2}{c|}{\multirow{2}{*}{25}} &
  \multicolumn{2}{c}{\multirow{2}{*}{0.3}} &
  \multicolumn{2}{c}{\multirow{2}{*}{0.6}} &
  \multicolumn{2}{c|}{\multirow{2}{*}{0.6}} &
  \multirow{2}{*}{0.5} &
  \multirow{2}{*}{0.4} &
  \multirow{2}{*}{0.5} &
  \multirow{2}{*}{0.65} &
  \multirow{2}{*}{0.65} &
  \multirow{2}{*}{0.65} \\
\multicolumn{2}{c|}{} &
  \multicolumn{2}{c}{} &
  \multicolumn{2}{c}{} &
  \multicolumn{2}{c|}{} &
  \multicolumn{2}{c}{} &
  \multicolumn{2}{c}{} &
  \multicolumn{2}{c|}{} &
   &
   &
   &
   &
   &
   \\ \bottomrule
\end{tabular}%
}
\caption{The detailed generation parameters for VariReal. We introduce the parameters for feasible and infeasible settings of three dataset respectively.}
\label{tab: app_gen}
\end{table*}
\begin{table*}[!htbp]
\centering
\resizebox{1.0\textwidth}{!}{%
\begin{tabular}{cl|ccccccc}
\toprule
\multicolumn{2}{c|}{\multirow{2}{*}{\textbf{HyperParameters}}} &
  \multirow{2}{*}{\textbf{lamda}} &
  \multirow{2}{*}{\textbf{lr}} &
  \multirow{2}{*}{\textbf{Min\_lr}} &
  \multirow{2}{*}{\textbf{Weight decay}} &
  \multirow{2}{*}{\textbf{Warm up steps}} &
  \multirow{2}{*}{\textbf{CLIP LoRA rank}} &
  \multirow{2}{*}{\textbf{CLIP LoRA alpha}} \\
\multicolumn{2}{c|}{} &  &  &  &  &       &       &      \\ \midrule
\multicolumn{2}{c|}{\multirow{2}{*}{Values}} &
  \multirow{2}{*}{0.5} &
  \multirow{2}{*}{\{1e-3,5e-4,1e-4,5e-5,1e-5\}} &
  \multirow{2}{*}{1e-08} &
  \multirow{2}{*}{1e-3, 1e-4, 5e-5} &
  \multirow{2}{*}{5\% total iterations} &
  \multirow{2}{*}{16} &
  \multirow{2}{*}{32} \\
\multicolumn{2}{c|}{} &  &  &  &  &       &       &      \\ \midrule
\multicolumn{2}{c|}{\multirow{2}{*}{\textbf{HyperParameters}}} &
  \multirow{2}{*}{\textbf{Training bs}} &
  \multirow{2}{*}{\textbf{Test bs}} &
  \multirow{2}{*}{\textbf{Train iterations}} &
  \multirow{2}{*}{\textbf{Val iterations}} &
  \multicolumn{3}{c}{\multirow{2}{*}{\textbf{Data augmentation}}} \\
\multicolumn{2}{c|}{} &  &  &  &  & \multicolumn{3}{c}{} \\ \midrule
\multicolumn{2}{c|}{\multirow{2}{*}{Values}} &
  \multirow{2}{*}{64} &
  \multirow{2}{*}{8} &
  \multirow{2}{*}{Pets:20700/AirC:72000/Cars:91840} &
  \multirow{2}{*}{1/70 Train iterations} &
  \multicolumn{3}{c}{\multirow{2}{*}{\begin{tabular}[c]{@{}c@{}}random resized crop, random horizontal flip, random color jitter, and\\      random gray scale\end{tabular}}} \\
\multicolumn{2}{c|}{} &  &  &  &  & \multicolumn{3}{c}{} \\  \bottomrule
\end{tabular}%
}
\caption{The hyper-parameter details for CLIP~\cite{clip} model fine-tuning.}
\label{tab: app_clip}
\end{table*}

We provide additional implementation details for VariReal in Table~\ref{tab: app_gen}. Key parameters include noise strength for the SDXL Inpainting model~\cite{sdinpaint} and conditioning strength for IP-Adapter~\cite{ip-adapter} with ControlNet~\cite{controlnet}. Due to varying difficulty across datasets and between feasible and infeasible generation, we use dataset-specific settings.

Following DataDream~\cite{kim2024datadream}, we tune learning rates and weight decay for classification tasks. We use a batch size of 64, AdamW~\cite{adam} optimizer, and a cosine annealing scheduler. Table~\ref{tab: app_clip} lists the CLIP~\cite{clip} fine-tuning parameters. Learning rates and weight decay are selected from a predefined range based on validation performance. The number of training iterations is fixed as described in \cref{sec:dataset}, with dataset-specific counts provided in the table.

\section{Other Image Editing Method Setups}
\label{app_others_detail}
As shown in Figure~\ref{fig: teaser}, we compare VariReal with InstructPix2Pix~\cite{instructpix2pix} and FPE~\cite{fpe}. To ensure fairness and leverage each model’s strengths, we follow their original usage guidelines. For FPE, we maintain aspect ratio via resizing and padding, and use the original training setup with recommended prompts—e.g., \texttt{"a [CLS] in the [ATTRIBUTE] background"} for background changes and \texttt{"a [ATTRIBUTE] [CLS]"} for color or texture edits, where [CLS] denotes the class name and [ATTRIBUTE] refers to feasible or infeasible prompts from \cref{sec: promptgen}. InstructPix2Pix uses prompts like \texttt{"put it in [ATTRIBUTE] background"} for background changes and \texttt{"make it a [ATTRIBUTE] aircraft"} for foreground edits. We conducted multiple trials and selected the best outputs for comparison.

\section{Method Details}
\label{sec: our_details}
\subsection{Guidance prompt}
\label{sec:guidance_prompt}

\begin{figure}[h!]
\centering
\includegraphics[width=1\linewidth]{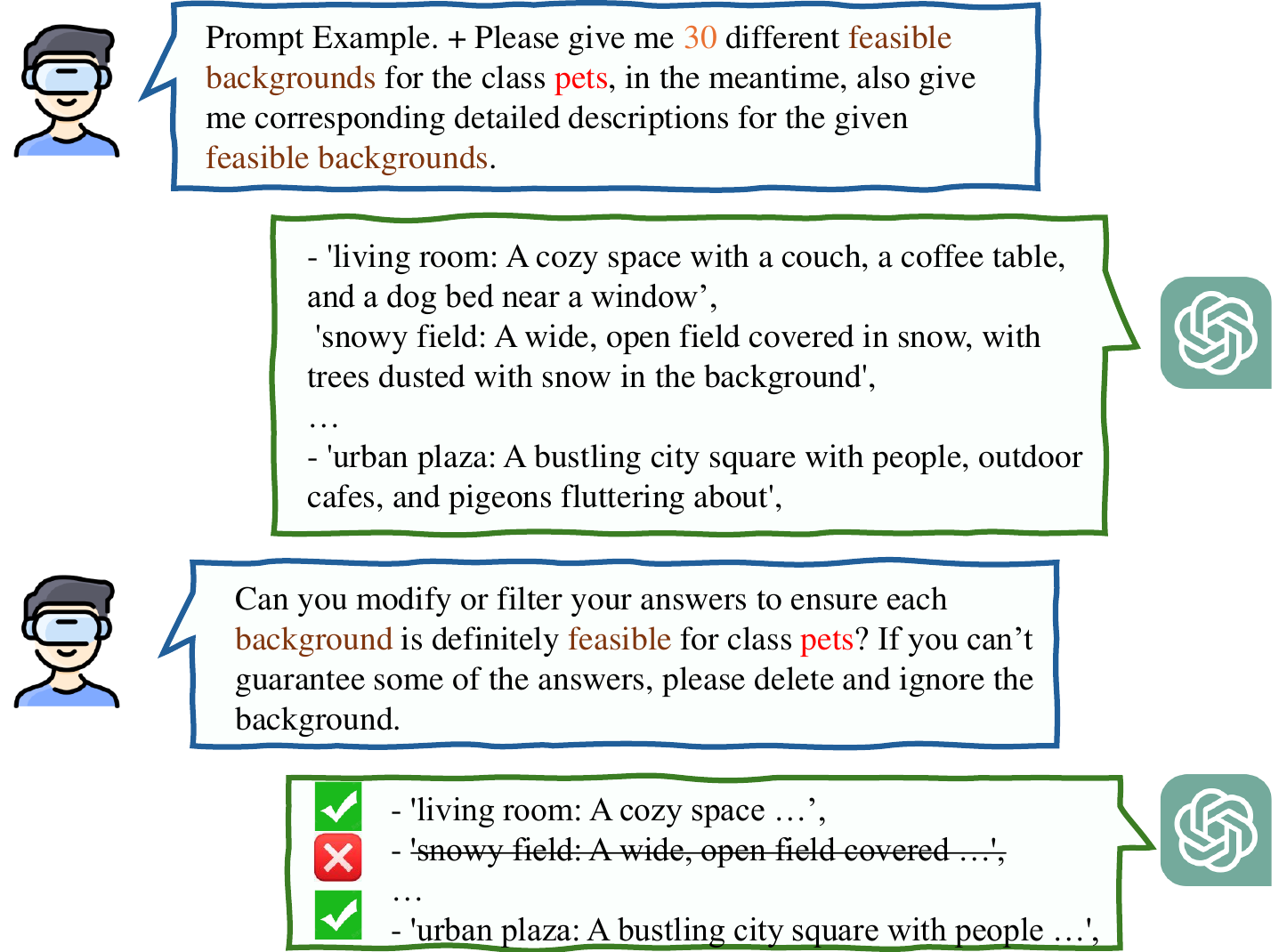}
\vspace{-1.0em}
\caption{The generated attributes(prompt words) and self-filtering process using ChatGPT-4~\cite{gpt4}.}
\label{gpt_gen}
\vspace{-1em}
\end{figure}

As detailed in \cref{sec: promptgen} and shown in Figure~\ref{gpt_gen}, the prompt generation process includes initial prompt generation and preliminary checks. 

Specifically, we use GPT-4~\cite{gpt4} to generate feasible or infeasible initial attributes (prompt words), which are then combined into a final prompt using our template: \( \text{"a photo of a [CLS]"} \), as shown in Figure~\ref{gpt_gen}. These initial attributes are then preliminarily checked by:

\begin{quote}
\raggedright
\texttt{"Can you modify or filter your answers to ensure each [background/color/texture] is definitely [feasible/infeasible] for class [CLASS]? Please delete and ignore some of the answers if you can't guarantee them."}
\end{quote}

For example, "deep cave" is not a feasible background for the pets class in the initial generation results and is filtered out by GPT-4. To ensure feasible attributes align with the training set, we manually check the existing backgrounds, colors, and textures in the training data and remove those absent from it. Table~\ref{tab: main1} shows the acceptance ratio at each stage. 

An example of generated attributes is the following, where the placeholders [ATTRIBUTE] represents the feasible/infeasible background, color, or texture, and [CLASS] represents a specific class.

\newpage  

\begin{mdframed}[backgroundcolor=white, linecolor=black, linewidth=0.5mm]
\textbf{Prompt Example.} \textit{"Task: As an AI language model, generate [Attribute] where the given class of objects typically exists ('feasible') and where they absolutely cannot exist ('unfeasible'). For each [Attribute], provide a one-sentence description detailing its visual appearance. You should adhere to the specified criteria.}

\textbf{Criteria:}
\begin{enumerate}
    \item Unique [Attribute]: Ensure each listed [Attribute] is distinct and not synonymous with others provided.
    \item Empty List Handling: If no unfeasible backgrounds can be identified, use 'EMPTY' to denote this.
    \item Format Requirement: Answers must be formatted as a Python list, following the structure shown in the 'Answer' section of the 'Example'.
\end{enumerate}

\textbf{Positive Example:}
\begin{itemize}
    \item \textbf{Object Class:} [CLASS]
    \item \textbf{Question:} Provide five different [Attribute] for the object class, each accompanied by a concise visual description.
    \item \textbf{Answer:}
    \begin{itemize}
        \item ...
    \end{itemize}
\end{itemize}

\textbf{Negative Examples:}
\begin{itemize}
    \item The answers are not acceptable as follows:
    \begin{itemize}
        \item ...
    \end{itemize}
    \item \textbf{Reasons:} ...
\end{itemize}

\textbf{Question:} Please give me [NUMBER] different [Attribute] for the class [CLASS]; in the meantime, also give me corresponding detailed descriptions for the given [Attribute].
\end{mdframed}

\begin{figure*}[ht!]
\centering
\includegraphics[width=1\linewidth]{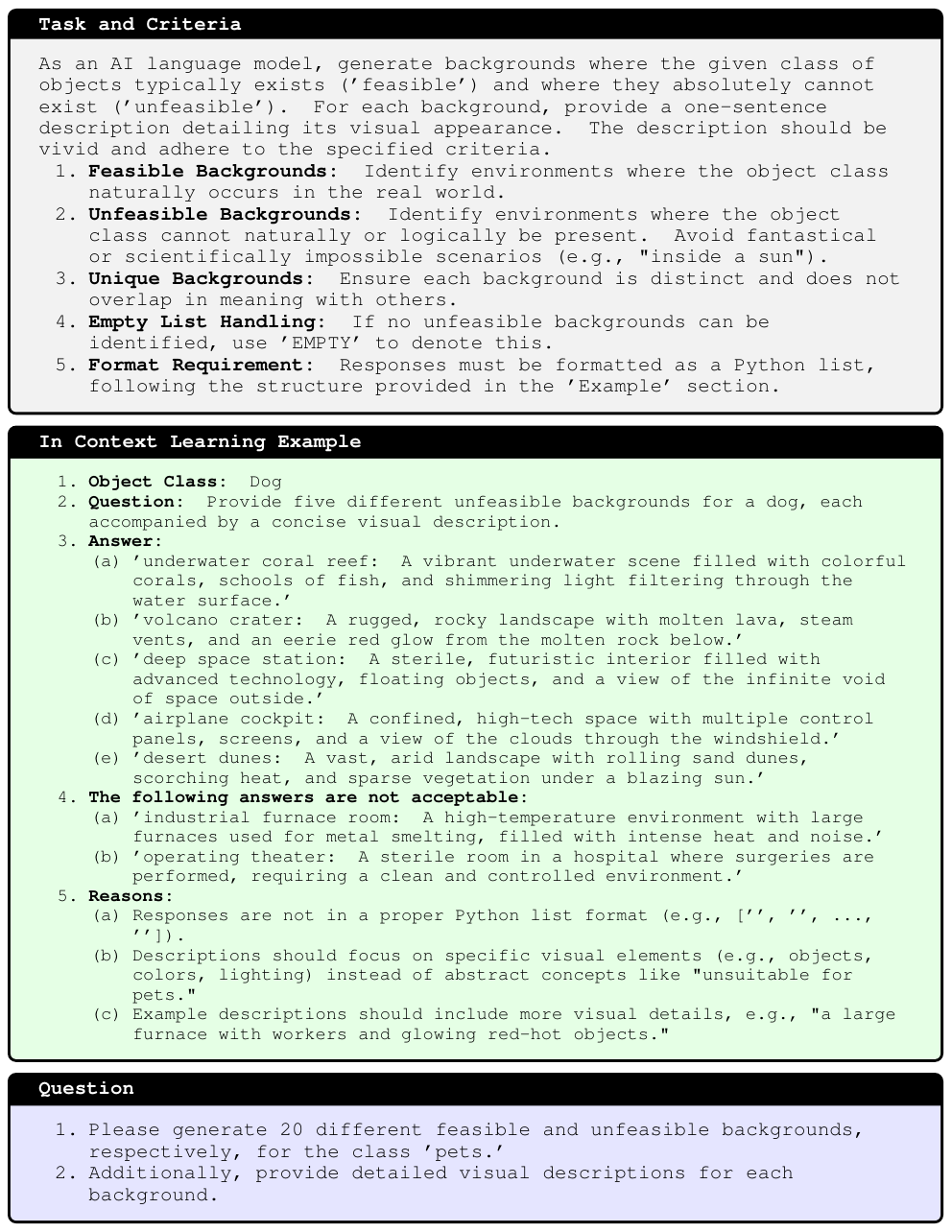}
\vspace{-2.0em}
\caption{A specific prompt example used for background prompt words generation of Oxford Pets~\cite{pets} dataset.}
\label{fig；app_fullprompt}
\vspace{-1em}
\end{figure*}

Here we also give one specific example for generating feasible and infeasible background for Oxford Pets dataset~\cite{pets} after replacing the placeholders in the above template in Figure~\ref{fig；app_fullprompt}.

\begin{figure*}[h!]
\centering
\includegraphics[width=1\linewidth]{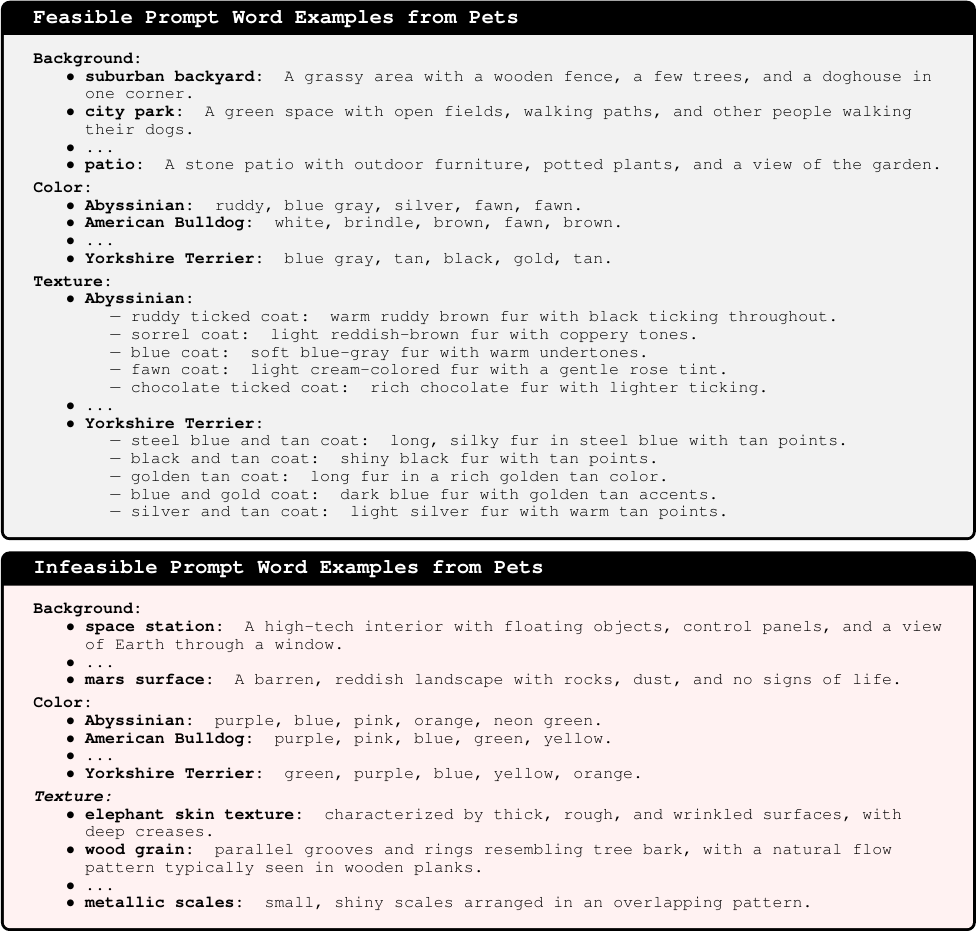}
\vspace{-1.0em}
\caption{Final accepted prompt word examples for Oxford Pets~\cite{pets}.}
\label{fig: app_promptwords}
\vspace{-1em}
\end{figure*}

By using the prompts described above, we also select some generated attributes (prompt words) to replace the placeholder in the prompt template. Due to space limitations, we provide up to five attributes as an example for the Oxford Pets~\cite{pets} dataset. Some generated feasible and infeasible prompt words can be found in Figure~\ref{fig: app_promptwords}.

\begin{figure*}[h!]
\centering
\includegraphics[width=0.8\linewidth]{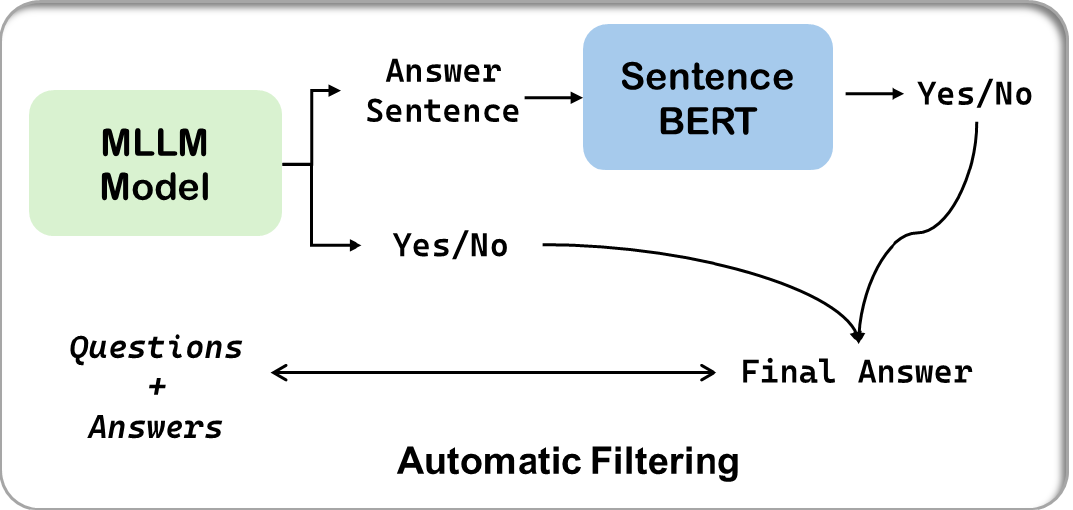}
\vspace{-1.0em}
\caption{The automatic filtering process using a MLLM model to filter the generated images using pre-defined qustions to check certain aspect for the generated image and ground truth answers.}
\label{filtering}
\vspace{-1em}
\end{figure*}

\begin{figure*}[h!]
\centering
\includegraphics[width=1\linewidth]{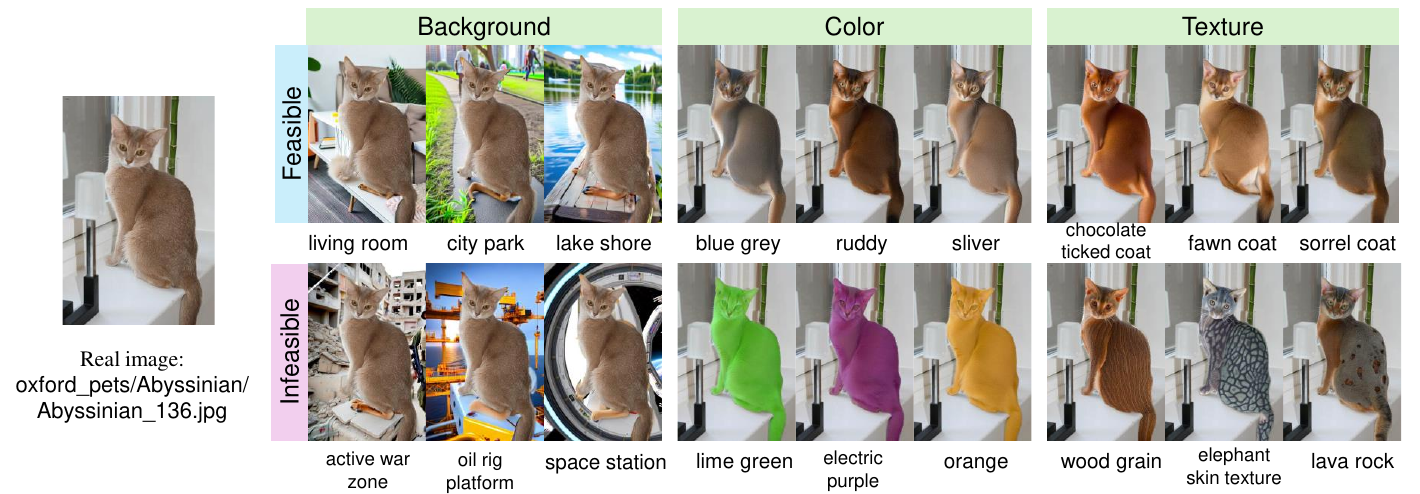}
\vspace{-1.0em}
\caption{Qualitative results of the class Abyssinian from Oxford Pets dataset~\cite{pets}, as a supplement for Figure~\ref{sample_1}.}
\label{oxpets_appendix}
\vspace{-1em}
\end{figure*}

\begin{figure}[h!]
\centering
\includegraphics[width=1\linewidth]{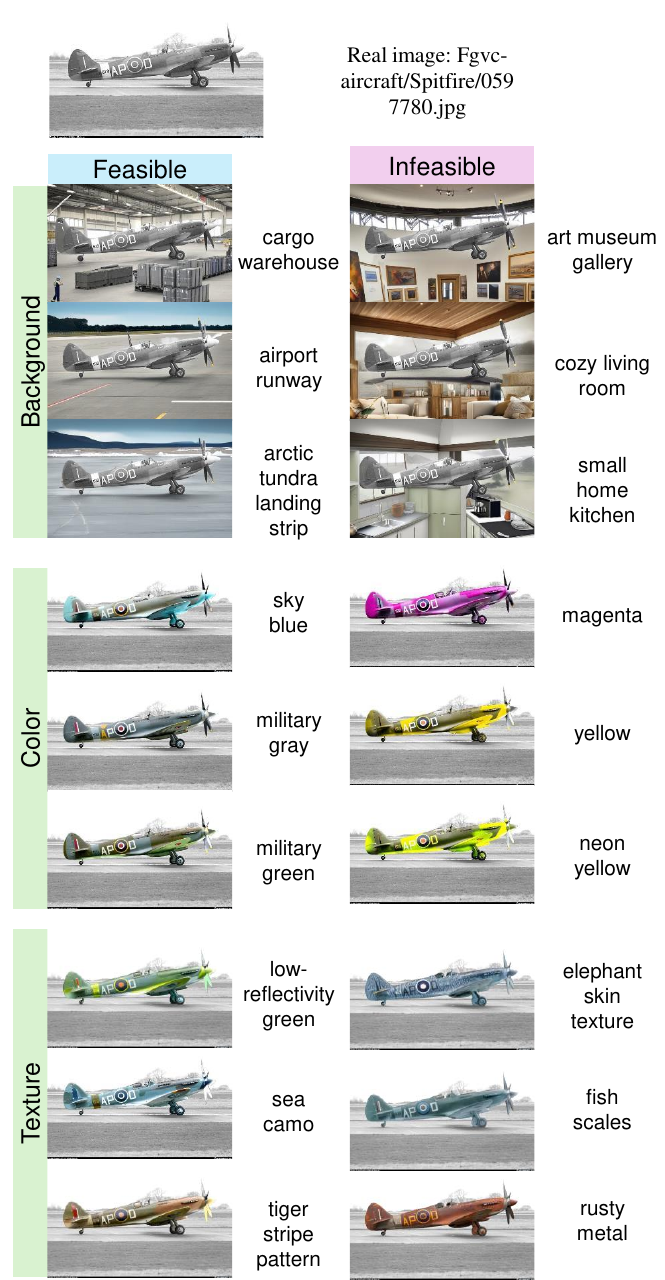}
\vspace{-1.0em}
\caption{Qualitative results of the class Spitfire from Fgvc-Aircraft dataset~\cite{aircraft}, as a supplement for Figure~\ref{sample_1}.}
\label{aircraft_appendix}
\vspace{-1em}
\end{figure}

\begin{figure}[h!]
\centering
\includegraphics[width=1\linewidth]{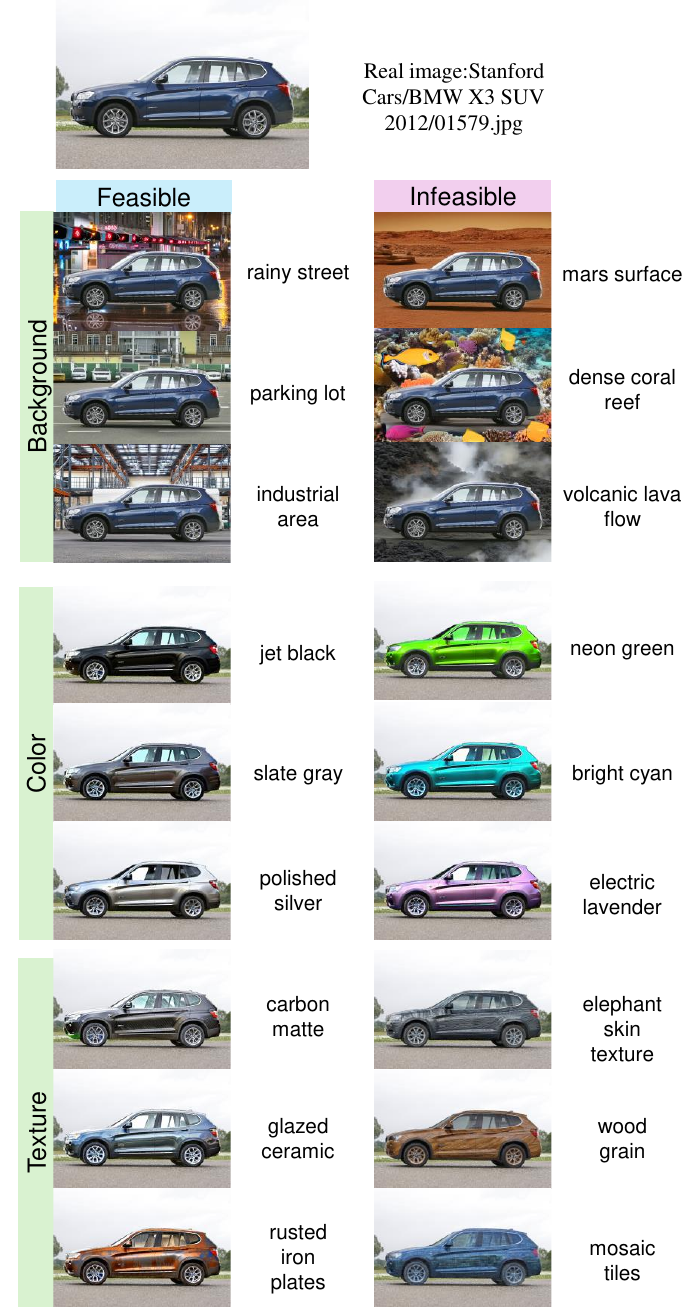}
\vspace{-1.0em}
\caption{TQualitative results of the class BMW X3 SUV 2012 from Stanford Cars dataset~\cite{cars}, as a supplement for Figure~\ref{sample_1}.}
\label{cars_appendix}
\vspace{-1em}
\end{figure}

\subsection{Automatic filtering}
\label{sec: app_af}

As introduced in \cref{sec: filtering}, we present the filtering questions for background, color, and texture changes. These checks ensure that the generated attributes align with the text prompt. For background attributes, we also verify if the foreground objects are feasible within the given background. Using placeholders for each background, color, texture prompt, object class, and feasibility information, we formulate questions based on the following filtering question template.

\begin{quote}
\raggedright
\textbf{Background-related questions:}
\begin{itemize}
    \item \textbf{Question 1:} Is the object in the image located in the [BACKGROUND] environment?  
    \textit{Choices:} ['yes', 'no']  
    \textit{Answer:} 'yes'  
    \item \textbf{Question 2:} Does the image background represent [BACKGROUND]?  
    \textit{Choices:} ['yes', 'no']  
    \textit{Answer:} 'yes'  
    \item \textbf{Question 3:} Does the [BACKGROUND] look feasible for the [CLS]?  
    \textit{Choices:} ['yes', 'no']  
    \textit{Answer:} 'yes' if [FEASIBLE] else 'no'  
    \item \textbf{Question 4:} Is it possible for the [CLS] in this image to exist in the real world with its background?  
    \textit{Choices:} ['yes', 'no']  
    \textit{Answer:} 'yes' if [FEASIBLE] else 'no'  
\end{itemize}
\textit{Note:} The placeholder [CLS] represents the current class name, [BACKGROUND] represents the target background being generated, and [FEASIBLE] denotes its feasibility.  
\end{quote}

If we change the color and texture, we use the following questions:

\textbf{Color and Texture-related questions:}
\begin{quote}
\raggedright
\begin{itemize}
    \item \textbf{Question 1:} Does the image show a [COLOR/TEXTURE] [CLS]?  
    \textit{Choices:} ['yes', 'no']  
    \textit{Answer:} 'yes'  
    \item \textbf{Question 2:} Is the [COLOR/TEXTURE] feasible for the [CLS]?  
    \textit{Choices:} ['yes', 'no']  
    \textit{Answer:} 'yes' if [FEASIBLE] else 'no'  
\end{itemize}
\textit{Note:} The placeholders retain similar meanings as above, where [COLOR/TEXTURE] indicates the current target appearance being generated.  
\end{quote}

\begin{figure}[h!]
\centering
\includegraphics[width=0.9\linewidth]{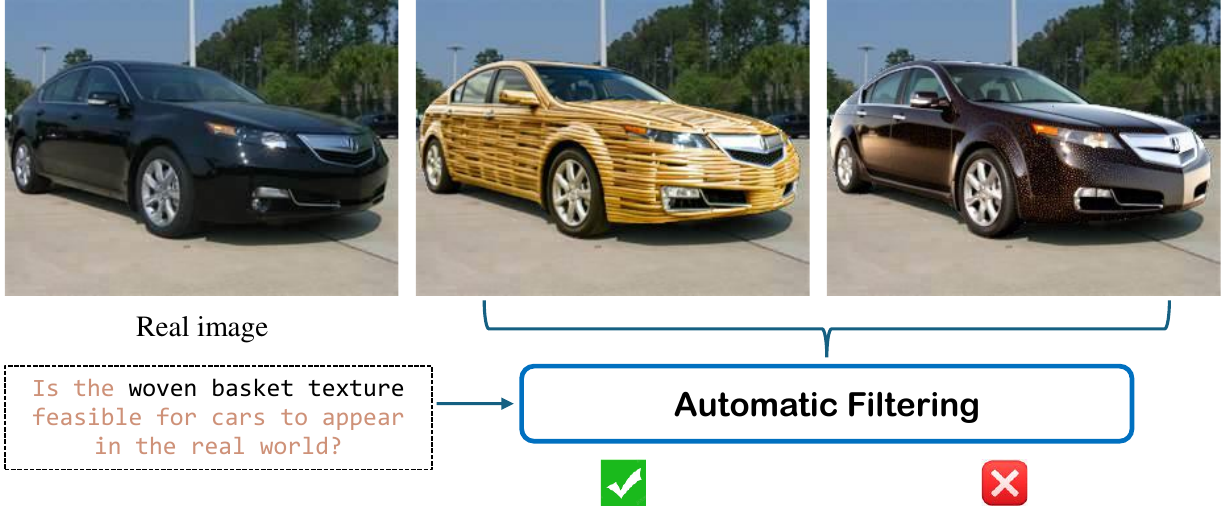}
\vspace{-1.0em}
\caption{Example of automatic texture filtering on the Cars~\cite{cars} dataset.}
\label{fig: app_automaticfiltering}
\vspace{-1em}
\end{figure}

We show an example process for the automatic filtering in the Figure~\ref{fig: app_automaticfiltering}.

\section{WaterBird Experiment Details}
\label{sec: app_waterbird}
In this section, we present detailed experimental results for the WaterBird~\cite{dunlap2023diversify} dataset under background modification settings, as shown in Table~\ref{tab: main3}. Notably, infeasible background edits improve performance by 5.8 percentage points in the synthetic-only setting and 1.6 percentage points in the real + synthetic setting.

\section{Classification Results Analysis}
\label{sec: app_classanyalysis}

In \cref{sec: ex_train_all_data}, we analyze mixing the feasible and infeasible data has no clear impact on classification tasks but some times will help the model learn complementary knowledge. We evaluate prediction correctness per test sample to compare knowledge learned by models trained under different settings. To measure whether one model's correctly predicted set is a subset of another's, we use: $\text{Inclusion Coefficient} = \frac{|A \cap B|}{|A|}$, with values closer to 1 indicating greater overlap. Additionally, we quantify the overlap of correctly predicted samples between models using the Jaccard index:$J(A, B) = \frac{|A \cap B|}{|A \cup B|}$, where \( A \) and \( B \), where \( A \) and \( B \) represent correct predictions from two training configurations.

\begin{figure*}[t]
\centering
\includegraphics[width=1.0\linewidth]{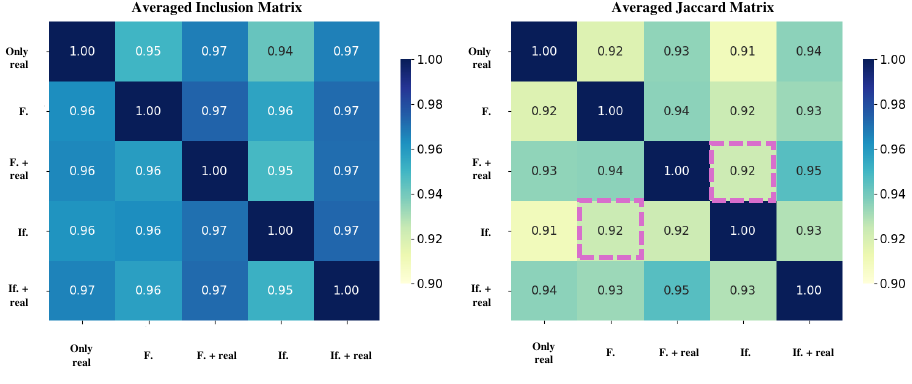}
\vspace{-1.8em}
\caption{The averaged Inclusion and Jaccard index matrix for three editing settings across three datasets. "f" = feasible, "if" = infeasible, "real" = training with real images. }
\label{learn_content}
\vspace{-1em}
\end{figure*}

The Inclusion matrix in Figure~\ref{learn_content} shows no subset relationship exists between model predictions.Notably, the feasible-only and infeasible-only settings labeled with dashed lines yield the lowest Jaccard scores, indicating minimal similarity.

\begin{observationbox}
\textit{Observation: The feasible and infeasible data lead the model to  learn different directions, while they achieve very similar performance.}
\end{observationbox}

\section{Qualitative Examples and User Study}
\label{sec: app_quaex}
We provide additional qualitative examples to demonstrate the generation quality of our VariReal method. One additional example from the Oxford Pets~\cite{pets}, FGVC Aircraft~\cite{aircraft}, and Stanford Cars~\cite{cars} datasets is included, along with one randomly selected example across these datasets.

\begin{figure*}[h!]
\centering
\includegraphics[width=0.9\linewidth]{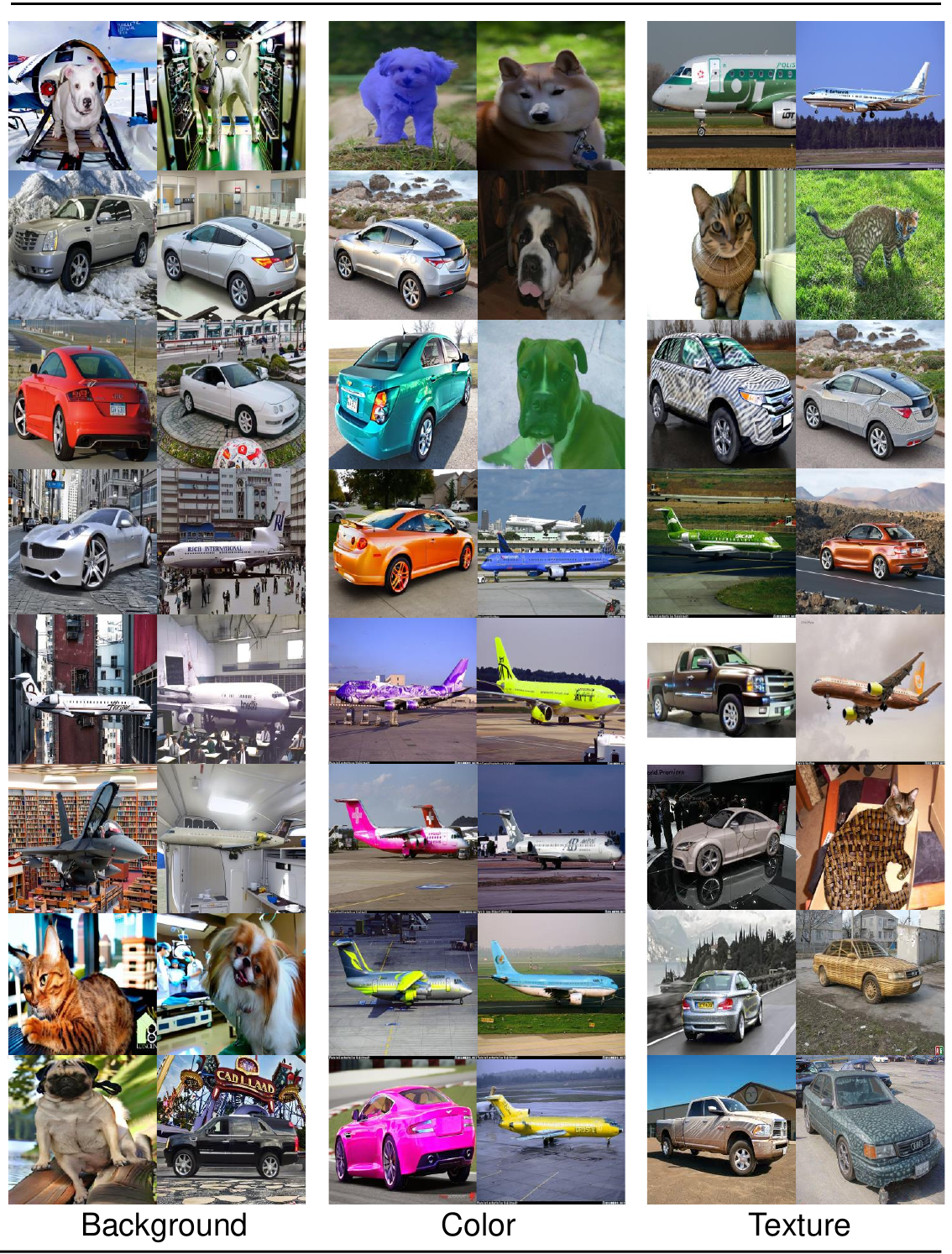}
\vspace{-1.0em}
\caption{Randomly selected generated samples across three datasets and feasibility attributes are shown. For visualization purposes, all images are resized to the same dimensions.}
\label{random}
\vspace{-1em}
\end{figure*}

\begin{figure*}[h!]
\centering
\includegraphics[width=0.8\linewidth]{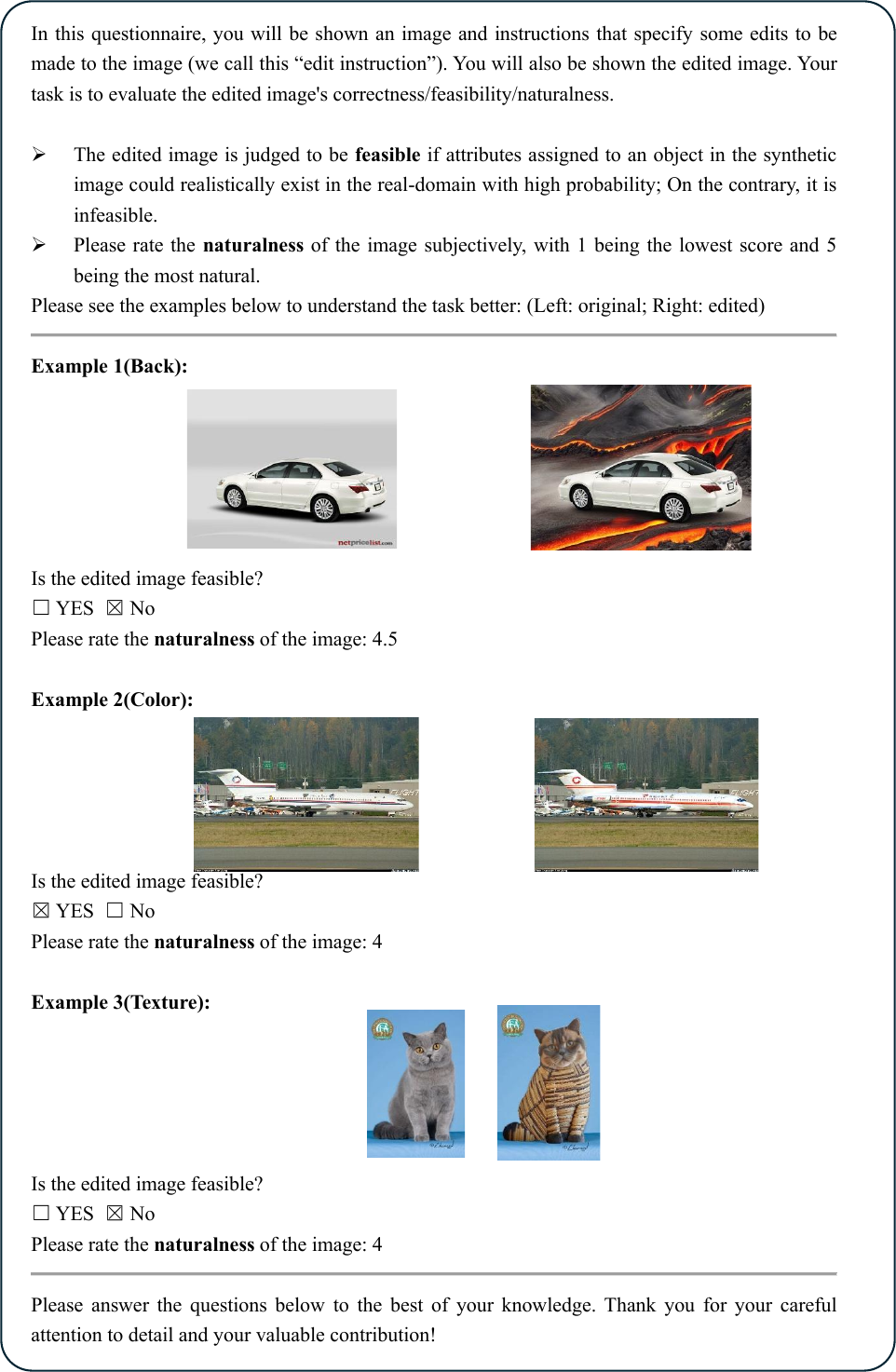}
\vspace{-1.0em}
\caption{Instructions for feasibility and naturalness generated images human study.}
\label{fig: app_question}
\vspace{-1em}
\end{figure*}

Figure~\ref{oxpets_appendix} shows the Abyssinian pet generation results, where our VariReal method produces more detailed backgrounds, such as "active war zone." Figure~\ref{aircraft_appendix} presents a Spitfire aircraft sample, illustrating snow in the background "arctic tundra landing strip." Figure~\ref{cars_appendix} features a BMW X3 SUV 2012 example. Finally, Figure~\ref{random} provides randomly selected examples from the three datasets for further visualization. The instruction for the questionnaire is shown in Figure~\ref{fig: app_question}.

\begin{figure*}[h!]
\centering
\includegraphics[width=0.8\linewidth]{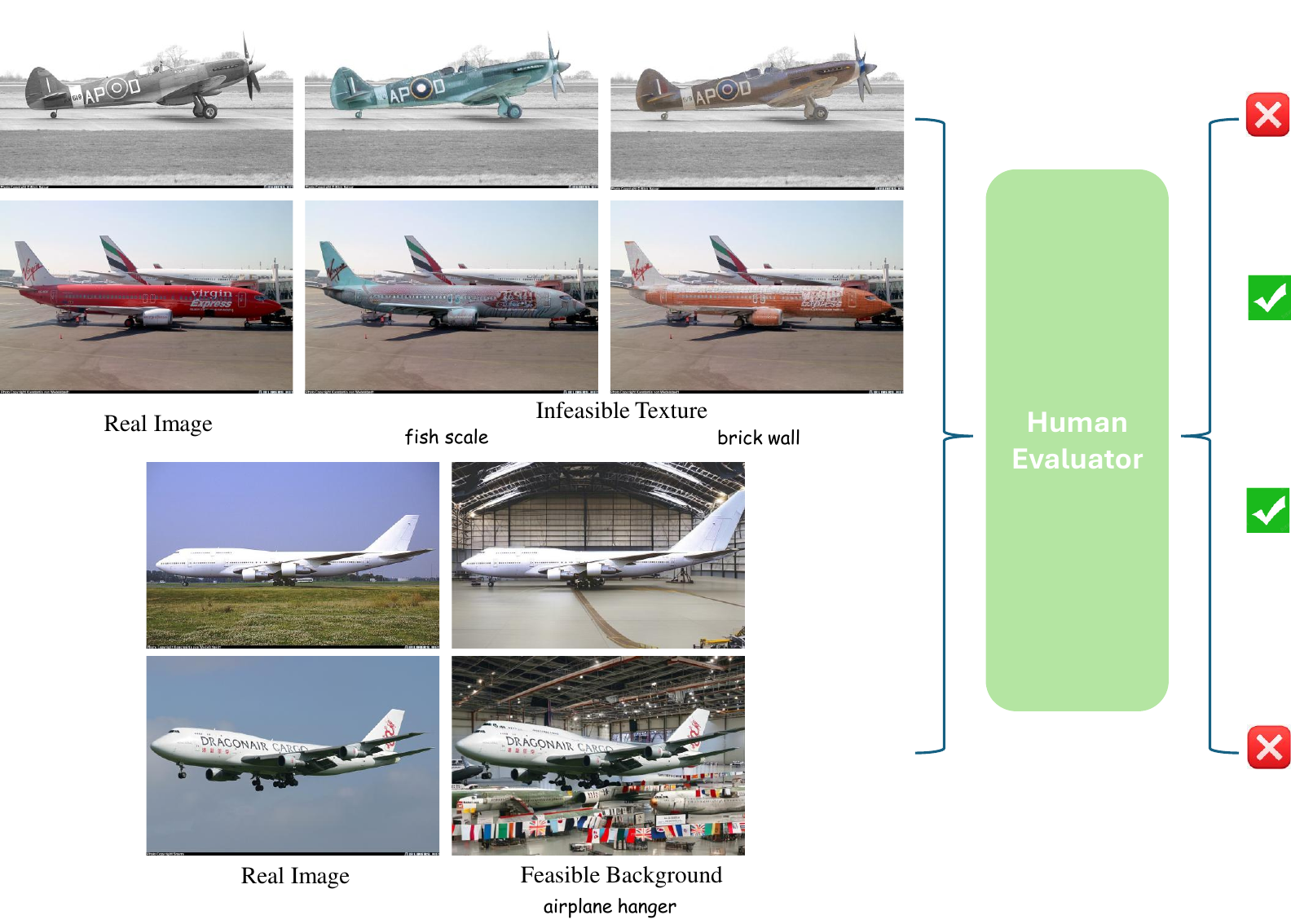}
\vspace{-1.0em}
\caption{Examples assessed as incorrect feasibility by human evaluators, including unclear fine-grained textures (e.g., "fish scale") and implausible object-background combinations (e.g., a flying aircraft inside a hangar).}
\label{fig: humanstudy_1}
\vspace{-1em}
\end{figure*}

Figure~\ref{fig: humanstudy_1} presents examples of correctly and incorrectly classified feasibility cases. More detail can be seen by zooming into the figures. For infeasible texture modifications, failure cases often like infeasible texture change of \textit{fish scale} or \textit{brick wall}, which are fine-grained and hard to represent clearly. In such cases, the output may only reflect the color rather than the intended texture, so human evaluators will classify these to the incorrect cases. Another source of error involves implausible object-background combinations—for example, a "flying aircraft in an airplane hangar" shown in the lower part of Figure~\ref{fig: humanstudy_1}.

\begin{figure*}[h!]
\centering
\includegraphics[width=0.8\linewidth]{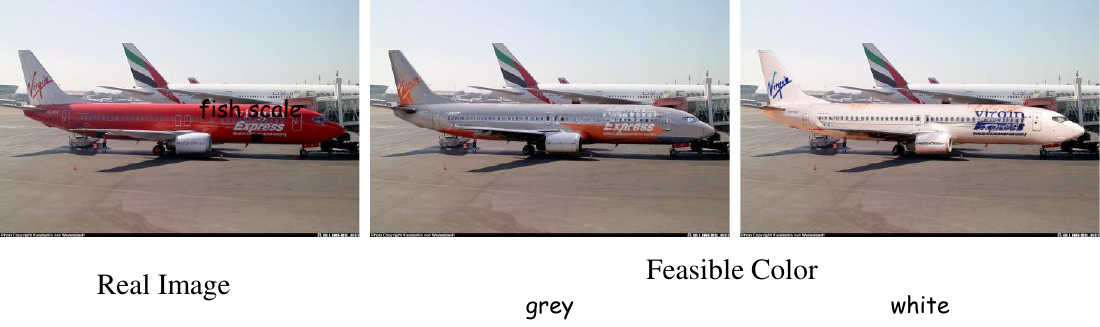}
\vspace{-1.0em}
\caption{Examples assessed by human evaluators as having lower naturalness, often due to unnatural color modifications or unrealistic visual appearance.}
\label{fig: humanstudy_2}
\vspace{-1em}
\end{figure*}

For the naturalness criterion, some images—such as those in Figure~\ref{fig: humanstudy_2} where the feasible color is changed from red to gray or white—receive lower scores, as the resulting colors appear less natural.

\begin{table}[h]
\centering
\renewcommand{\arraystretch}{0.8} 
\footnotesize 
\resizebox{0.7\columnwidth}{1.5cm}{%
\begin{tabular}{c|cc|cc}
\toprule
 &
  \textbf{R} &
  \textbf{S} &
  \multicolumn{2}{c}{\textbf{WaterBirds~\cite{dunlap2023diversify}}} \\
   & & & F & IF \\  
\midrule
{} 0-shot &
   &
   &
  \multicolumn{2}{c}{79.0} \\ 
\midrule
\multirow{2}{*}{Back.} &  
   &
   \ding{51} &
  86.6 & 
  92.4 \\
 &  
   \ding{51} &
   \ding{51} &
  92.9 & 
  94.5 \\ 
\midrule
{} Real &
   \ding{51} & 
   &
  \multicolumn{2}{c}{85.7} \\ 
\bottomrule
\end{tabular}%
}
\vspace{-0.5em}
\caption{The top-1 performance using the full training set and synthetic data, with training setups including synthetic-only and synth. + real data. The attribute of experimented dataset WaterBirds~\cite{dunlap2023diversify} is background. All results use synthetic images set to five times the number of real images. }
\label{tab: main3}
\vspace{-1.0em}
\end{table}

\section{Ablation Study}
\label{sec: app_ablation}

\begin{figure*}[h!]
\centering
\includegraphics[width=1.0\linewidth]{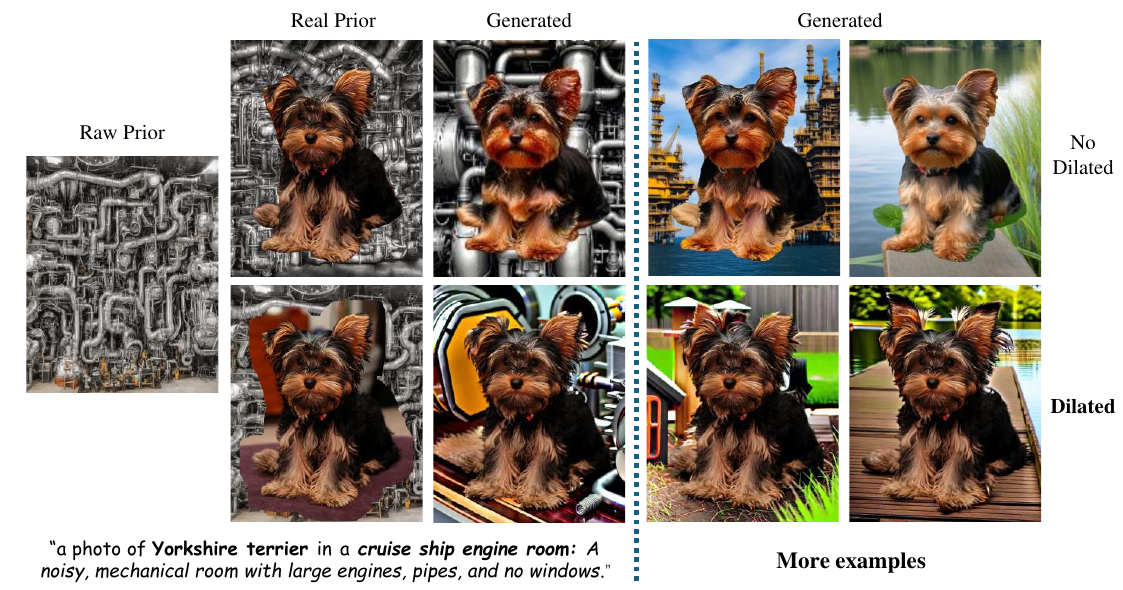}
\vspace{-1.0em}
\caption{The ablation study for the usage to expand object mask for background edition setting. We show the real generated prior background on the left, and then present the different combined image with real and prior image.}
\label{back_ab}
\vspace{-1em}
\end{figure*}

We ablate the mask dilation step introduced in \cref{sec: prior}, which helps maintain spatial coherence between objects and backgrounds. Without mask dilation, generated images often exhibit a "floating" effect shown in Figure~\ref{back_ab}, where objects appear unnaturally integrated into their backgrounds.

\end{document}